\documentclass[lettersize,journal]{IEEEtran}
\usepackage{amsmath,amsfonts}
\usepackage{algorithmic}
\usepackage{algorithm}
\usepackage{array}
\usepackage[caption=false,font=normalsize,labelfont=sf,textfont=sf]{subfig}
\usepackage{textcomp}
\usepackage{stfloats}
\usepackage{url}
\usepackage{verbatim}
\usepackage{graphicx}
\usepackage{cite}
\usepackage{balance}
\usepackage{bbm}
\newcounter{ALG@line}
\begin{document}

\title{ConsRoute: Consistency-Aware Adaptive Query Routing for Cloud–Edge–Device Large Language Models}

\author{Haoyu~Qiao, Hao~Zhang, Shanwen~Mao, Siyao~Cheng, and Jie~Liu,~\IEEEmembership{Fellow,~IEEE}
\thanks{This work is partly supported by the Project of Laboratory of Advanced Agricultural Sciences of Heilongjiang Province under Grant No. ZY04JD05-010,  the Key Research and Development Program of Heilongjiang Province under Grant No. 2022ZX01A22, 
and the National Natural Science Foundation of Heilongjiang Province under Grant No. YQ2019F007. (Corresponding author:
Hao Zhang.)}
\thanks{Haoyu~Qiao, Zhang Hao, Shanwen~Mao, and Siyao Cheng are with the Department of Computer Science and Technology, Harbin Institute of Technology,
Harbin 150001, China and also with the National Key Laboratory of Smart Farm Technologies and Systems, Harbin 150001, China (e-mail:24s103393@stu.hit.edu.cn; zhh1000@hit.edu.cn; 24s103313@stu.hit.edu.cn;csy@hit.edu.cn).}
\thanks{Jie Liu is with the Department of Computer Science and Technology, Harbin Institute of Technology (ShenZhen), ShenZhen 518055, China and also with the National Key Laboratory of Smart Farm Technologies and Systems, Harbin 150001, China (e-mail: jieliu@hit.edu.cn).}}

\markboth{Journal of \LaTeX\ Class Files,~Vol.~14, No.~8, August~2021}%
{Shell \MakeLowercase{\textit{et al.}}: A Sample Article Using IEEEtran.cls for IEEE Journals}

\IEEEpubid{0000--0000/00\$00.00~\copyright~2021 IEEE}

\maketitle

\begin{abstract}

Large language models (LLMs) deliver impressive capabilities but incur substantial inference latency and cost, which hinders their deployment in latency-sensitive and resource-constrained scenarios. Cloud–edge–device collaborative inference has emerged as a promising paradigm by dynamically routing queries to models of different capacities across tiers. In this paper, we propose ConsRoute, a lightweight, semantic-aware, and adaptive routing framework that significantly improves inference efficiency while minimizing impact on response quality.
Unlike prior routing methods that rely on predicting coarse-grained output quality gaps, ConsRoute leverages a reranker to directly assess the semantic consistency between responses generated by models at different tiers, yielding fine-grained soft supervision signals for routing. To minimize device-side overhead, ConsRoute reuses hidden states from the LLM prefilling stage as compact query representations, avoiding additional encoders or inference passes. Furthermore, these representations are clustered, and Bayesian optimization is employed to learn cluster-specific routing thresholds that dynamically balance quality, latency, and cost under heterogeneous query distributions.
Extensive experiments demonstrate that ConsRoute achieves near-cloud performance ($\geq$95\%) while reducing end-to-end latency and inference cost by nearly 40\%, consistently outperforming existing routing baselines in both response quality and system efficiency.
\end{abstract}

\begin{IEEEkeywords}
Large Language Model Inference, Cloud–Edge–Device Collaboration, Adaptive Query Routing, Semantic Consistency, Latency and Cost Optimization.
\end{IEEEkeywords}

\section{Introduction}

\IEEEPARstart{L}{arge} language models (LLMs) have demonstrated remarkable capabilities across a wide spectrum of tasks, including natural language understanding, reasoning, and generation~\cite{brown2020language, bai2023qwen, touvron2023llama}. These advances have spurred growing interest in deploying LLM-powered services in mobile and ubiquitous computing environments, such as personal assistants, intelligent sensing, and edge intelligence applications. However, bringing LLMs into such resource-constrained settings exposes a fundamental and increasingly critical tension: while larger models generally offer superior response quality and robustness, they incur prohibitive inference latency, computation, and energy costs; conversely, smaller models are significantly more efficient but often suffer from degraded reasoning and generation quality.

This trade-off is particularly pronounced in mobile and IoT systems, where device-side computation, memory capacity and battery lifetime are all limited and dynamically varying. As a result, no single LLM instance can simultaneously satisfy the diverse and often conflicting requirements of real-world mobile applications, including high response quality, low end-to-end latency, and low device-side resource consumption. Addressing this challenge is therefore of both practical importance and fundamental relevance to mobile computing systems, motivating the design of collaborative inference frameworks that coordinate multiple LLMs across the device, edge, and cloud to jointly optimize quality, latency, and cost.

\begin{figure}[!t]
    \centering
    \includegraphics[width=\columnwidth]{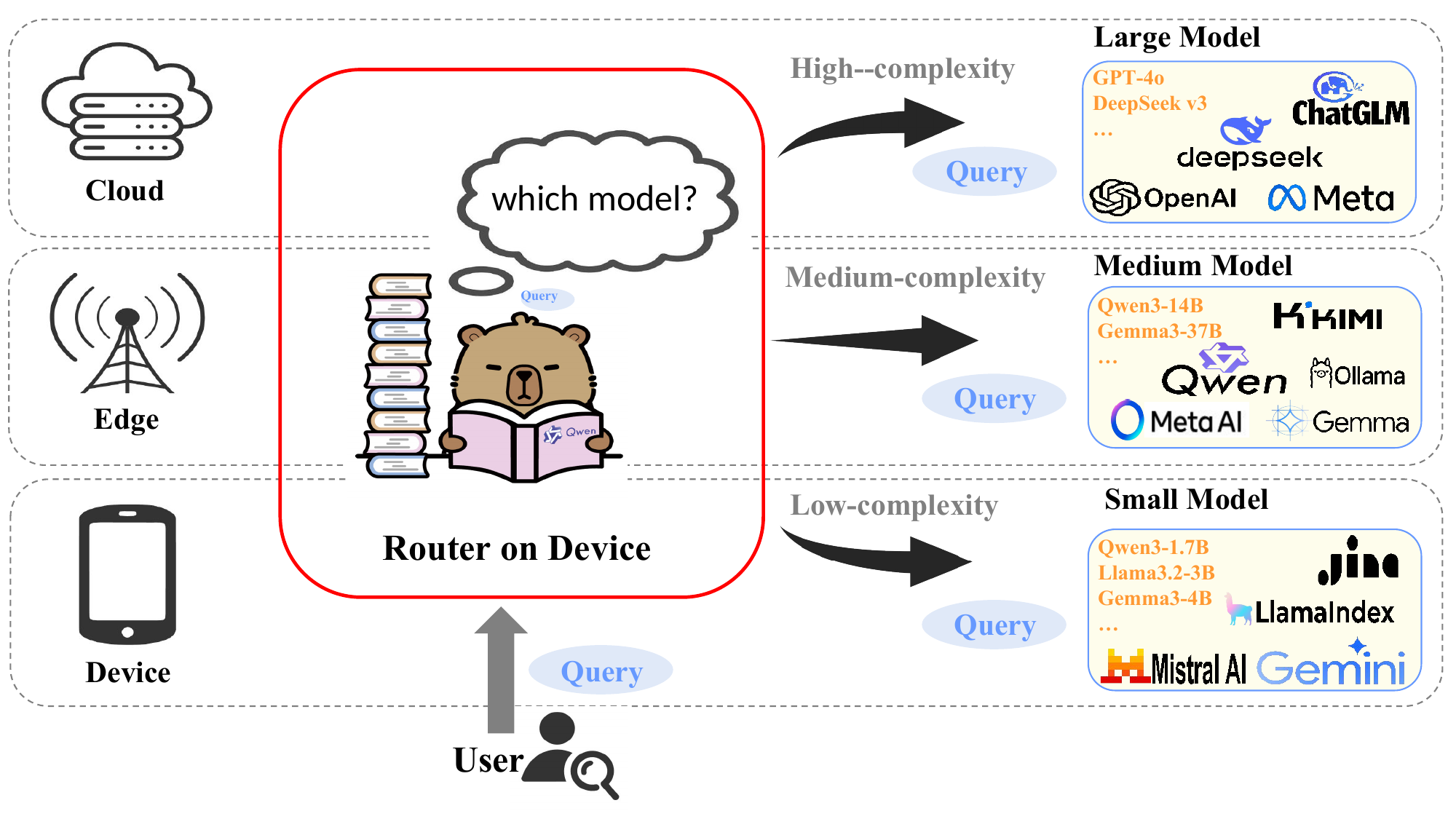} 
    \caption{Cloud-Edge-Device LLM Collaboration System}
    \label{fig:cloud-edge-device}
\end{figure}
\IEEEpubidadjcol
A widely adopted paradigm to realize such collaboration is hierarchical deployment across the cloud--edge--device continuum~\cite{zhang2024edgeshard, qu2025mobile}. As illustrated in Figure~\ref{fig:cloud-edge-device}, a large-scale LLM is deployed in the cloud (CLM), e.g., 30B\,+ parameter models such as DeepSeek-V3~\cite{liu2024deepseek} or Qwen3-32B~\cite{yang2025qwen3}, providing high-quality responses for complex or high-risk queries at the expense of higher latency and communication overhead. At the edge, a medium-scale LLM (ELM), typically in the 7--30B range (e.g., Qwen3-14B~\cite{yang2025qwen3}), serves as a middle ground, offering a more favorable balance between capability and cost. On the device, a lightweight LLM (DLM), often with fewer than 4B parameters (e.g., Qwen3-1.7B~\cite{yang2025qwen3}), is co-located with a query router to provide low-latency and energy-efficient responses for simple or low-risk requests. By selectively invoking different model tiers, this architecture enables adaptive trade-offs between performance and efficiency in dynamic mobile environments.

At the core of this hierarchical architecture lies query routing, which determines whether a given request should be processed locally on the device, offloaded to the edge, or escalated to the cloud~\cite{varangot2025doing}\cite{kassem2025robust}. An effective routing decision hinges on whether the response generated by a lower-tier model is sufficiently consistent with that of a more capable model. If high consistency can be anticipated, the system can safely rely on the device-side model, thereby minimizing latency and resource usage; otherwise, the query should be forwarded to a higher-tier model to ensure response quality. Consequently, accurately predicting cross-model response consistency before generation is a key enabler for efficient and reliable collaborative inference.

Early work, such as FrugalGPT~\cite{frugalgpt} and AutoMix~\cite{automix}, adopts a cascading strategy that sequentially invokes models from smallest to largest, escalating only when the confidence of the smaller model’s output falls below a predefined threshold. While conceptually simple, this approach incurs additional latency due to the need to generate full responses at multiple levels. More recent methods, including Zooter~\cite{zooter}, RouteLLM~\cite{ong2024routellm}, and OptLLM~\cite{liu2024optllm}, attempt to mitigate this overhead by making routing decisions prior to generation, typically via learned predictors that estimate the expected quality or utility of candidate models. These predictive routing approaches enable faster model selection and have demonstrated improved efficiency.

Despite these advances, existing routing methods still suffer from several fundamental limitations. 
First, many approaches rely on offline-obtained scalar quality or reward scores and use the difference between small and large models as supervision for router training. Mapping rich, structured text outputs to a single scalar inevitably discards fine-grained semantic information and fails to capture nuanced inconsistencies between model responses. 
Second, query representations are often extracted using fine-tuned BERT-style encoders or external embedding APIs, which either impose substantial on-device computation overhead or introduce additional communication latency, making them ill-suited for real-time mobile scenarios. 
Third, most existing methods employ a single, global threshold to determine routing decisions. Such static thresholds cannot adapt to heterogeneous query types, task sensitivities, or evolving data distributions, often resulting in suboptimal trade-offs between quality and efficiency.

To address these challenges, we propose \textbf{ConsRoute}, a lightweight, consistency-aware query routing framework for cloud--edge--device collaborative LLM inference. ConsRoute advances the state of the art in three key aspects. First, inspired by re-ranking techniques in information retrieval, we directly measure the semantic gap between model outputs in a high-dimensional space, constructing routing supervision signals based on textual semantic similarity rather than indirect scalar rewards. This provides a more faithful characterization of cross-model consistency. Second, we design a lightweight consistency predictor that reuses the DLM’s prefill hidden states, avoiding additional text encoding. By appending a short prompt after the query and extracting the final-layer hidden state of the last token, we obtain a compact semantic representation that is fed into a small MLP for routing, significantly reducing on-device latency and resource consumption. Third, we introduce a dynamic thresholding mechanism based on query clustering and Bayesian optimization, enabling routing policies to adapt to different query categories and system conditions in an online and data-driven manner.

Our contributions are summarized as follows:
\begin{itemize}
    \item We propose {ConsRoute}, a consistency-aware query routing mechanism for cloud--edge--device collaborative LLM inference, which leverages semantic similarity between model outputs and efficiently reuses DLM prefill hidden states to enable fast and lightweight on-device routing.
    \item We design a dynamic, online-updatable thresholding strategy based on query clustering and Bayesian optimization, allowing routing decisions to adapt to heterogeneous query types and evolving workloads.
    \item We integrate ConsRoute into a hierarchical collaborative inference system and demonstrate through extensive experiments that it achieves response quality comparable to large cloud models, while reducing end-to-end latency and inference cost by nearly 40\%.
\end{itemize}

The remainder of this paper is organized as follows. Section~II reviews related work on cloud--edge--device collaborative LLM inference and multi-LLM query routing. Section~III formalizes the cloud--edge--device collaborative inference setting, including the hierarchical architecture, routing policy, and optimization objective. Section~IV presents the proposed ConsRoute framework in detail, covering prompt-guided semantic representation extraction, semantic-consistency supervision, and cluster-based adaptive thresholding with online Bayesian optimization. Section~V reports extensive experimental results on RouterBench and additional benchmarks, together with ablation studies, online adaptation analysis, and network condition sensitivity. Section~VI concludes the paper and discusses future research directions.

\section{Related Work}

\begin{table*}[t]
    \caption{Comparison of representative multi-LLM methods. 
    Columns indicate whether a method targets cloud--edge--device deployment (For Cloud-Edge-Device Deployment), 
    performs pre-generation model choice (Predictive Routing), 
    avoids additional query encoders beyond the LLMs (No-Extra Encoder), 
    reuses device-side LLM internal states for routing (Reuse DLM info), 
    employs non-global or task-aware thresholds (Adaptive Threshold), 
    and supports online updating of the routing policy (Online Router updating). 
    $\checkmark$ denotes support, $\times$ denotes lack of support, and ``--'' denotes not applicable.}
    \label{tab:rw-summary}
    \centering
    \footnotesize
    \setlength{\tabcolsep}{6pt}%
    \renewcommand{\arraystretch}{1.15}%
    \begin{tabular*}{\textwidth}{@{\extracolsep{\fill}}lcccccc}
        \hline
        Method &
        For Cloud-Edge-Device &
        Predictive &
        No-Extra &
        Reuse &
        Adaptive &
        Online \\
        & Deployment & Routing & Encoder & DLM info & Threshold & Router updating \\
        \hline
        VELO~\cite{yao2025enhancing}        & $\checkmark$ & $\checkmark$ & $\checkmark$ & $\times$      & --      & $\checkmark$ \\
        Mixture of Thought~\cite{yue2023large} & $\times$    & $\times$    & $\checkmark$    & $\checkmark$ & -- & $\times$    \\
        FrugalGPT~\cite{frugalgpt}          & $\times$      & $\times$    & $\times$ & $\times$ & $\times$ & $\times$    \\
        AutoMix~\cite{automix}              & $\times$      & $\times$    & $\checkmark$ & $\checkmark$ & -- & $\times$    \\
        HybridLLM~\cite{hybridllm}          & $\times$      & $\checkmark$ & $\times$ & $\times$    & $\times$ & $\times$    \\
        Zooter~\cite{zooter}                & $\times$      & $\checkmark$ & $\checkmark$ & $\times$    & -- & $\times$    \\
        RouteLLM~\cite{ong2024routellm}     & $\times$      & $\checkmark$ & $\times$ & $\times$    & $\times$ & $\times$    \\
        ME\mbox{-}Switch~\cite{liu2024meswitch} & $\times$   & $\checkmark$ & $\times$ & $\times$    & -- & $\times$    \\
        MixLLM~\cite{wang2025mixllm}        & $\times$      & $\checkmark$ & $\times$ & $\times$    & --      & $\checkmark$ \\
        \textbf{ConsRoute (ours)}           & $\checkmark$ & $\checkmark$ & $\checkmark$     & $\checkmark$ & $\checkmark$      & $\checkmark$ \\
        \hline
    \end{tabular*}
\end{table*}

\subsection{Cloud--Edge--Device Collaborative LLM Inference}

Collaborative inference across cloud, edge and device has become a key paradigm for deploying large models in mobile and ubiquitous computing scenarios~\cite{9937169}\cite{li2025collaborative}. These systems aim to jointly optimize end-to-end latency, resource utilization, and quality of service (QoS) by carefully deciding where to execute each request under constrained computation, bandwidth, and energy budgets. Instead of running a single monolithic model in the cloud, they exploit the heterogeneous capabilities of cloud servers, edge nodes, and end devices to deliver more responsive and cost-effective services.

He et al.~\cite{cloud-edge-offload} propose an active-inference-based offloading framework for LLM tasks in cloud--edge environments, addressing the data inefficiency, latency insensitivity, and poor adaptability to workload shifts observed in prior deep reinforcement learning solutions. VELO~\cite{yao2025enhancing} introduces a vector-database-assisted framework that leverages multi-agent reinforcement learning (MARL) to optimize QoS for edge LLM users, reducing both latency and resource consumption while significantly improving user satisfaction. Hao et al.~\cite{hao2024hybrid} proposes a dynamic token-level Edge-Cloud collaboration for LLM Inference. These works exemplify how intelligent scheduling and resource allocation can substantially improve the performance of LLM services in distributed infrastructures. CE-LSLM~\cite{zhu2025lslm} introduced a key-value (KV) cache reuse mechanism to enhance the semantic understanding of edge models through contextual guidance from the cloud, while significantly reducing edge-side computational and storage overhead

\subsection{Query Routing for Multi-LLM Systems}

Beyond macro-level task offloading and resource management, an essential research direction in multi-LLM systems is query routing: deciding, for each incoming request, which model tier to invoke under accuracy, latency, and cost constraints. In hierarchical cloud--edge--device architectures, the router is the core component that bridges user queries and heterogeneous models, and its behavior directly determines the overall efficiency and reliability of the system. Existing work explores a variety of routing paradigms, including cascaded inference strategies that escalate from small to large models, learned routers trained under different supervision signals derived from model outputs, and diverse architectural choices and online adaptation mechanisms. In the following, we review these lines of work.

\subsubsection{Cascaded Inference Strategies}

Early approaches to multi-LLM collaboration often adopt cascaded inference strategies that sequentially invoke models of increasing capacity and cost. The central idea is to start from a cheaper model, estimate its confidence in the generated response, and only escalate to a larger model when confidence is deemed insufficient. This design aims to trade off quality and cost without requiring complex predictors or upfront routing models.

Mixture of Thought~\cite{yue2023large} treats the self-consistency of multiple samples from a weaker LLM as a signal of problem difficulty, proposing several sampling and consistency-checking schemes. While this improves over naive single-pass decoding, it still relies on the model’s own confidence and does not address overconfidence in small models; moreover, sampling multiple full-length responses significantly increases latency. FrugalGPT~\cite{frugalgpt} arranges multiple LLMs into a cost-ordered chain and uses a DistilBERT classifier~\cite{sanh2019distilbert} to predict the correctness of each intermediate answer; if the predicted correctness is low, the router escalates to the next, more expensive model. AutoMix~\cite{automix} models the escalation process as a partially observable Markov decision process (POMDP), allowing a small model to evaluate the confidence of its own outputs and decide whether to invoke a higher-tier model.

Although cascaded strategies can effectively reduce average cost under certain workloads, they are inherently non-predictive and sequential: the system must wait for the small model to generate (possibly multiple) responses before making a routing decision. This design introduces additional routing latency and is particularly problematic in mobile and edge settings, where end-to-end responsiveness and energy efficiency are critical. 

\subsubsection{Supervision Signals for Query Routing}

The choice of supervision signal is fundamental to the performance of a learned query router, as it determines which property of model behavior the router is trained to approximate. Most prior methods construct labels by evaluating and comparing the quality of outputs from different models, effectively using some notion of quality difference as the target for training.

HybridLLM~\cite{hybridllm} trains a DeBERTa-based router to predict the quality gap between a small language model (SLM) and a large language model (LLM), where the gap is defined as the difference in BartScore between their outputs. Zooter~\cite{zooter} uses a reward model (QwenRM~\cite{bai2023qwen}) to assign utility scores to model responses and distills it into a lightweight classifier that makes routing decisions based on expected utility; it further performs label augmentation by combining per-query rewards with cluster-level average rewards. RouteLLM~\cite{ong2024routellm} leverages human preference data from Chatbot Arena, training routers to estimate model preferences across queries using several learning strategies. In all these methods, the supervision is derived from scalar quality, preference, or reward scores, and routing labels are defined by comparing such scores across models.

The core idea behind these methods is to evaluate the quality of the outputs of different models and then use the difference in quality as a supervision signal. While effective in many scenarios, this approach of mapping rich textual information to a one-dimensional quality or reward score and then calculating the difference between them inevitably loses information about fine-grained semantic relationships between outputs and may fail to capture fine-grained semantic consistency between responses. 

\subsubsection{Router Architectures and Online Adaptation}

Beyond supervision signals, existing work also differs in how the query router itself is architected and how it adapts to changing workloads or constraints. A router must both understand query semantics sufficiently to estimate model performance and remain lightweight enough for deployment in resource-constrained mobile or edge environments. Achieving this balance between expressiveness and efficiency is a central design challenge in cloud--edge--device LLM systems.

OptLLM~\cite{liu2024optllm} formulates routing as a multi-objective optimization problem that jointly considers accuracy and cost, employing random forests and ensemble voting to estimate whether each LLM can produce a correct answer and to output confidence scores for different models. ME-Switch~\cite{liu2024meswitch} concatenates the query with a routing prompt and feeds it into a fine-tuned Qwen1.5-1.8B model, using the prompted router’s explicit output to select a model. MetaLLM~\cite{nguyen2024metallm} casts routing as a multi-armed bandit problem that adaptively balances performance and cost through online exploration and exploitation. MixLLM~\cite{wang2025mixllm} introduces label-augmented embeddings and online optimization to dynamically control routing under latency constraints, improving robustness to workload variations. InferenceDynamics~\cite{shi2025inferencedynamics} proposes a flexible and scalable multi-dimensional routing framework that explicitly models the capability and knowledge profiles of a large pool of specialized LLMs.

These methods highlight rich design spaces for router architectures and online adaptation, but they typically treat the router as a separate, often heavyweight module that does not reuse internal states of the underlying LLMs. This separation leads to significant additional computational and memory overhead, making it less suitable for on-device deployment and difficult to achieve a balance between accurate query understanding and lightweight implementation.

Table~\ref{tab:rw-summary} summarizes the characteristics of the most representative methods. We compare them with our proposed method ConsRoute, with detailed explanations provided in Appendix B.
The comparison highlights that ConsRoute is, to the best of our knowledge, the only approach that simultaneously targets cloud--edge--device deployment, avoids additional encoders by reusing DLM states, and supports adaptive, online-updated routing thresholds, making it particularly suitable for resource-constrained mobile settings.

\section{Problem Setting}

In this section, we formalize the cloud--edge--device collaborative inference problem considered in this work. We first describe the hierarchical deployment of LLMs across different tiers. We then define queries, responses, and the routing policy that determines where each query is served. Finally, we introduce the performance metrics of interest and formulate the routing objective.

\subsection{Hierarchical Cloud--Edge--Device Architecture}

We consider a hierarchical architecture composed of three tiers: the device tier, the edge tier, and the cloud tier. Each tier hosts a LLM with different capacity, latency, and resource requirements, referred to as the DLM, ELM, and CLM, respectively. The set of available serving options is denoted by
\[
    \mathcal{M} = \{\mathrm{DLM}, \mathrm{ELM}, \mathrm{CLM}\}.
\]

User queries are generated at the device tier. Upon the arrival of each query, the system can either process it locally using the DLM or offload it to the ELM or CLM for inference. We focus on the design of a routing policy that dynamically selects an appropriate serving tier from $\mathcal{M}$ for each incoming query, aiming to balance response quality, latency, and computational cost.

\subsection{Query, Response, and Routing Policy}

Let $\mathcal{X}$ denote the space of user queries, and let $x \in \mathcal{X}$ represent a query sampled from an underlying workload distribution. For each model $m \in \mathcal{M}$, we denote by $y_m(x)$ the textual response generated by model $m$ when serving query $x$.

A routing policy determines which tier is responsible for answering each query. Formally, we model the router as a policy
\[
    \pi: \mathcal{X} \rightarrow \mathcal{M},
\]
which maps a query $x$ to a selected serving option $\pi(x) \in \mathcal{M}$. Given a fixed routing policy $\pi$, the system behavior is fully determined: each query $x$ is always served by the model $m = \pi(x)$, incurring the corresponding quality, latency, and cost characteristics of that tier.

\subsection{Performance Metrics}

Under a routing policy $\pi$, the system induces a distribution over how queries are assigned to different model tiers. This distribution, in turn, determines the overall answer quality, end-to-end latency, and computational cost of the system.

We denote by $\mathrm{Acc}(\pi)$ the expected answer quality under policy $\pi$, measured as the average correctness of responses over all served queries. We denote by $\mathrm{Latency}(\pi)$ the expected end-to-end latency, which includes both computation time and communication delays associated with offloading. Finally, we denote by $\mathrm{Cost}(\pi)$ the expected inference cost under policy $\pi$. Following~\cite{wilkins2024cost}, we estimate $\mathrm{Cost}(\pi)$ based on the total number of activated model parameters multiplied by the number of generated tokens. Together, these metrics provide a compact yet expressive characterization of the trade-offs induced by different routing strategies.

\begin{figure*}[htbp]
    \centering
    \includegraphics[width=0.95\textwidth]{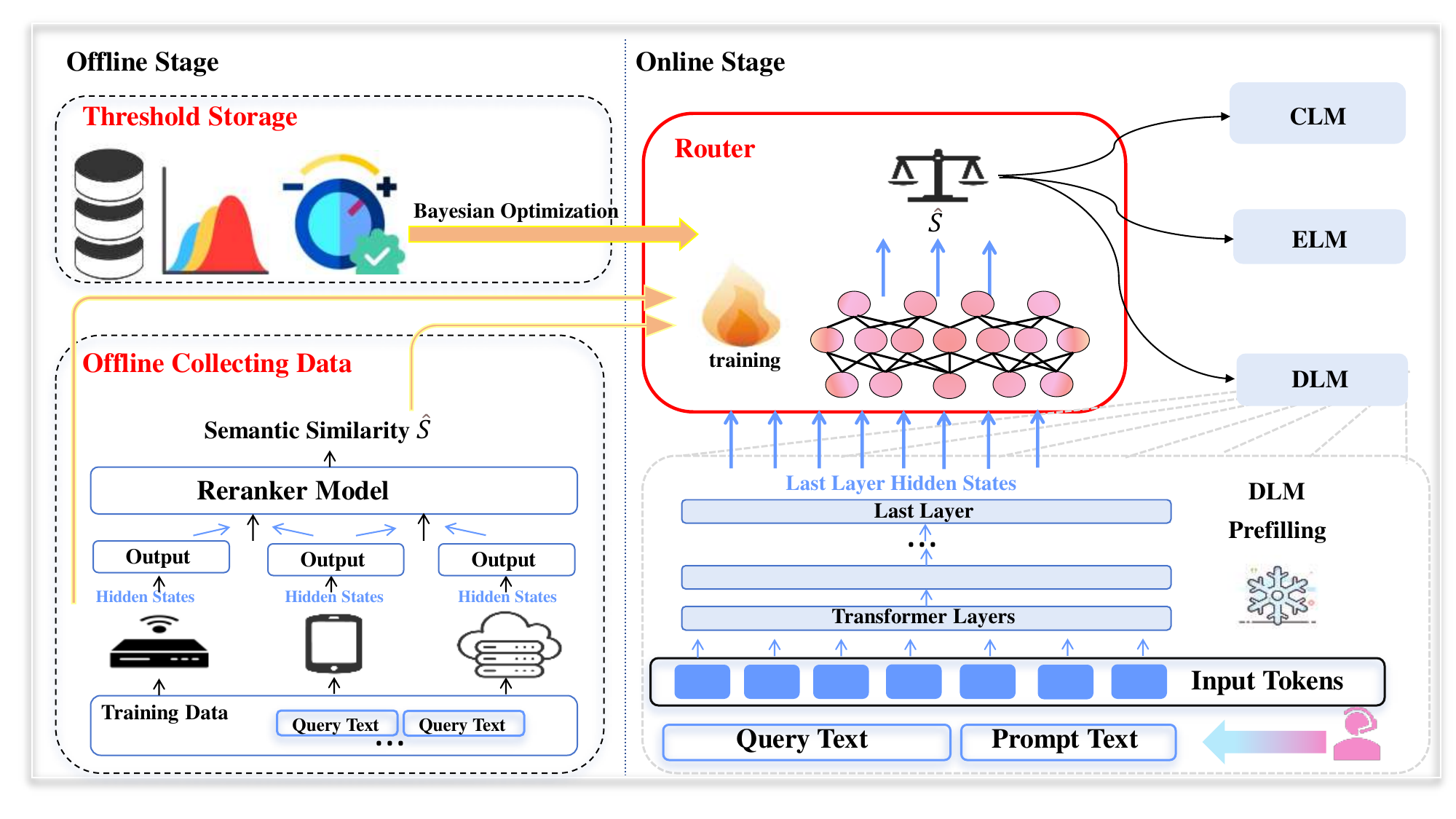}
    \caption{Overview of the ConsRoute framework. The bottom right shows the Semantic Representation Extractor, which leverages the DLM to extract input semantics (Section~\ref{Semantic Representation}). The bottom left shows the Training Data Construction process for the predictor and the top right shows the Lightweight Consistency Predictor, which decides which model tier a query should be routed to (Section~\ref{Lightweight Consistency Predictor}). The top left presents the Adaptive Routing Policy, where appropriate routing thresholds are determined via Bayesian optimization (Section~\ref{Adaptive Threshold via Bayesian Optimization}).}
    \label{fig:overview}
\end{figure*}

\subsection{Optimization Objective}

The objective of the routing policy is to preserve, as much as possible, the answer quality achieved by always invoking the strongest model (i.e., the CLM), while significantly reducing inference latency and computational cost by routing suitable queries to lower-tier, more efficient models. This naturally leads to a multi-objective optimization problem over the space of routing policies.

A convenient way to express this goal is through a scalar utility function
\[
    U(\pi)
    = f\big(\mathrm{Acc}(\pi), \mathrm{Latency}(\pi), \mathrm{Cost}(\pi)\big),
\]
where $f(\cdot)$ is a monotone function that increases with answer quality and decreases with latency and cost (e.g., a weighted combination of these three terms). The optimal routing policy is then defined as
\[
    \pi^\star \in \arg\max_{\pi} U(\pi).
\]

In the remainder of the paper, we instantiate this abstract formulation with a concrete, lightweight routing framework that leverages device-side signals to approximate favorable trade-offs between response quality, latency, and cost in practical cloud--edge--device inference settings.

\section{ConsRoute: Consistency-Aware Query Routing for Cloud--Edge--Device LLMs}

In this section, we first outline the overall workflow of our proposed consistency-guided query routing framework and summarize the key questions that drive its design. We then provide a detailed description of each component and answer these questions. Table~\ref{tab:notation} lists the main mathematical notations used in the subsequent sections.

\subsection{Overview of ConsRoute}

\begin{table}[!t]
\caption{Main Notations Used in Our Method.\label{tab:notation}}
\centering
\renewcommand{\arraystretch}{1.05}
\begin{tabular}{>{\centering\arraybackslash}m{0.14\linewidth}>{\centering\arraybackslash}m{0.7\linewidth}}
\hline
\textbf{Item}\rule{0pt}{2.6ex} & \textbf{Description}\rule{0pt}{2.6ex} \\
\hline
DLM & Large Language Model on the Device tier.\\[0.6ex]
ELM & Large Language Model on the Edge tier. \\[0.6ex]
CLM & Large Language Model on the Cloud tier. \\[0.6ex]
$x$ & User query.\\[0.6ex]
$p$ & Fixed natural-language instruction prompt.\\[0.6ex]
$x'$ & Concatenated input sequence $x' := x \,\|\, p \,\|\, \text{[EOS]}$.\\[0.6ex]
$h_T$ & Consistency-aware representation extracted from the DLM for $x'$.\\[0.6ex]
$f_{\text{rank}}$ & Pre-trained reranker used to compute semantic similarity between responses.\\[0.6ex]
$\mathrm{SIM}(x)$ & Reranker-based semantic similarity between DLM and stronger model outputs.\\[0.6ex]
$\mathrm{AUG}(x)$ & Augmented supervision signal from references or an LLM judge.\\[0.6ex]
$S_{\text{cloud}}(x)$ & Consistency label for the DLM--CLM response pair.\\[0.6ex]
$S_{\text{edge}}(x)$ & Consistency label for the DLM--ELM response pair.\\[0.6ex]
$S_{\text{fused}}(x)$ & Fused soft label combining $S_{\text{cloud}}(x)$ and $S_{\text{edge}}(x)$.\\[0.6ex]
$f_\theta$ & Lightweight MLP head that predicts consistency from $h_T$.\\[0.6ex]
$\hat{S}(x)$ & Predicted consistency score $\hat{S}(x) := f_\theta(h_T) \in [0,1]$.\\[0.6ex]
$\mathcal{C}_k$ & $k$-th semantic cluster of queries in representation space.\\[0.6ex]
$c_k$ & Centroid of cluster $\mathcal{C}_k$.\\[0.6ex]
$(\tau_1^{(k)}, \tau_2^{(k)})$ & Cluster-specific routing thresholds for DLM/ELM/CLM.\\[0.6ex]
$\mathcal{U}_k$ & Cluster-level utility that trades off accuracy, latency, and cost.\\[0.6ex]
$u(x)$ & Per-query utility sample used for online threshold adaptation.\\[0.6ex]
$\lambda_1, \lambda_2, \lambda_3$ & Weights balancing accuracy, latency, and cost in the utility.\\[0.6ex]
\hline
\end{tabular}
\end{table}

We propose \textbf{ConsRoute}, a lightweight consistency-aware query routing framework designed for hierarchical LLM deployment across the device, edge, and cloud tiers. The core idea is to route each query to the cheapest model whose response is expected to remain semantically consistent with that of a stronger model, thereby preserving answer quality while substantially reducing latency and inference cost.

As summarized in Figure~\ref{fig:overview}, ConsRoute operates in two phases: an offline supervision construction phase (left) and an online routing phase (right). In the offline phase, we construct semantically grounded consistency labels that directly measure the agreement between responses generated by models at different tiers, which are used to train a lightweight on-device predictor and to initialize cluster-specific routing thresholds. In the online phase, as detailed in Algorithm~\ref{alg:consistency-routing}, ConsRoute reuses the device-side LLM’s prefilling computation to extract a consistency-aware semantic representation of each query, predicts its expected consistency with stronger models, and applies adaptive, cluster-specific thresholds to determine whether the query should be served locally, escalated to the edge, or forwarded to the cloud.

This design allows ConsRoute to jointly address three key challenges in multi-tier LLM deployment: (i) enabling accurate routing decisions without incurring additional on-device inference overhead, (ii) supervising routing using semantic consistency rather than coarse scalar quality gaps, and (iii) adapting routing policies to heterogeneous query types and evolving system conditions.

Concretely, ConsRoute consists of four key components:

\begin{enumerate}
    \item \textbf{Consistency Label Construction:}
    Offline, we build soft consistency labels by comparing the device-side answer with the edge-side and cloud-side answers using a reranker model, and then lightly augment these similarity scores with correctness and alignment signals from references or an LLM judge. These labels are used as supervision for the router.

    \item \textbf{Prompt-Guided Representation Extraction:}
    On the device, the DLM processes the user query concatenated with a fixed instruction during the prefilling stage and produces a task-aware hidden representation at the last token. This representation is encouraged to encode how consistent the DLM’s answer would be with stronger models, while avoiding any extra encoder or additional forward pass.

    \item \textbf{Lightweight Consistency Prediction:}
    A compact MLP head is then trained to map the DLM representation to a predicted consistency score, using the offline consistency labels as regression targets. This yields an efficient on-device predictor that can estimate device–edge–cloud agreement without running reward models, rerankers, or external APIs at inference time.

    \item \textbf{Cluster-Based Adaptive Thresholding:}
    Finally, we cluster queries in the representation space and assign each cluster its own routing thresholds, which are tuned by Bayesian optimization over a utility function that jointly considers accuracy, latency, and cost. During deployment, these cluster-specific thresholds are further refined online as traffic patterns and network conditions evolve, enabling adaptive trade-offs between quality and efficiency.
\end{enumerate}

\begin{algorithm}[t]
\caption{Consistency-Guided Hierarchical Routing with Adaptive Thresholds}
\label{alg:consistency-routing}
\begin{algorithmic}[1]
\REQUIRE Query $x$, prompt $p$, frozen DLM, trained predictor $f_\theta$, cluster centroids $\{c_k\}_{k=1}^K$, thresholds $\{(\tau_1^{(k)}, \tau_2^{(k)})\}_{k=1}^K$ (learned by Algorithm~\ref{alg:adaptive-thresholds})
\ENSURE Selected model tier $\in \{\text{DLM}, \text{ELM}, \text{CLM}\}$

\STATE \refstepcounter{ALG@line}\label{line:construct-input} Construct input $x' \leftarrow x \,\|\, p \,\|\, [\text{EOS}]$
\STATE \refstepcounter{ALG@line}\label{line:encode} Extract semantics $h_T \leftarrow \text{Encode}(x'; \text{DLM})$
\STATE Predict routing score $\hat{S}(x) \leftarrow f_\theta(h_T)$
\STATE \refstepcounter{ALG@line}\label{line:assign-cluster} Assign cluster index $k^* \leftarrow \arg\min_k \|h_T - c_k\|_2$
\STATE \refstepcounter{ALG@line}\label{line:retrieve-thresholds} Retrieve thresholds $(\tau_1, \tau_2) \leftarrow (\tau_1^{(k^*)}, \tau_2^{(k^*)})$

\IF{$\hat{S}(x) > \tau_1$}
    \RETURN DLM
\ELSIF{$\hat{S}(x) > \tau_2$}
    \RETURN ELM
\ELSE
    \RETURN CLM
\ENDIF
\end{algorithmic}
\end{algorithm}

In the following, we detail each component of ConsRoute by answering the following research questions:

\begin{itemize}
    \item \textbf{RQ1:} 
    How can we extract a consistency-aware semantic representation for routing by reusing the device-side LLM, without introducing additional encoders or incurring noticeable on-device overhead?

    \item \textbf{RQ2:} 
    How can we construct reliable and semantically grounded supervision signals that accurately capture response consistency across device-, edge-, and cloud-level LLMs?

    \item \textbf{RQ3:} 
    How can we design routing thresholds that adapt to heterogeneous queries and evolving workloads, while achieving favorable trade-offs between answer quality, latency, and inference cost?

\end{itemize}

\subsection{Semantic Representation (RQ1)}
\label{Semantic Representation}
\paragraph{Motivation}
During the prefilling stage of generation, the DLM has already processed the full input query and computed deep hidden representations for all tokens. This provides a valuable opportunity to directly reuse these hidden states for semantic representation extraction. To make the representation task-aware, we lightly guide the DLM using a fixed natural language prompt appended to the query. 

\begin{figure}[!t]
    \centering
    \includegraphics[width=0.8\columnwidth]{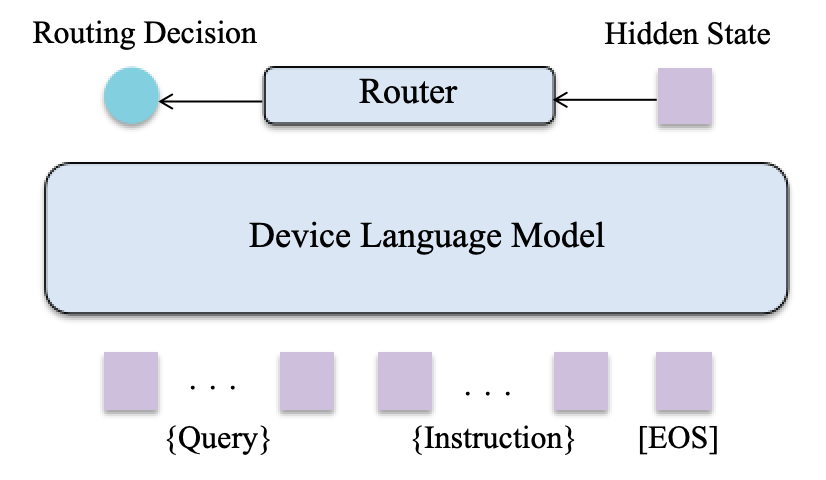} 
    \caption{Prompt-guided representation learning. The user query  is concatenated with a fixed instruction and an EOS token.The DLM processes the input and the final-layer hidden state of the EOS token  is used as a consistency-aware representation for routing.}
    \label{fig:prompt concat}
\end{figure}

\paragraph{Prompt-Guided Representation}
To obtain a consistency-aware representation without deploying an additional encoder, we let the DLM itself produce a task-specific embedding through a prompt, as illustrated in Figure~\ref{fig:prompt concat}.

Given a user query $x$, we first append a fixed natural-language instruction $p$ and then terminate the sequence with an end-of-sequence token [EOS]. The resulting input token sequence is
\begin{equation}
x' := x \,\|\, p \,\|\, \text{[EOS]},
\label{eq:prompt-input}
\end{equation}
where the instruction prompt is

\noindent\hspace{1em}\texttt{``The consistency between the small language model and large language model responses for the above query is:''}

\begin{figure*}[htbp]
    \centering
    \includegraphics[width=0.9\textwidth]{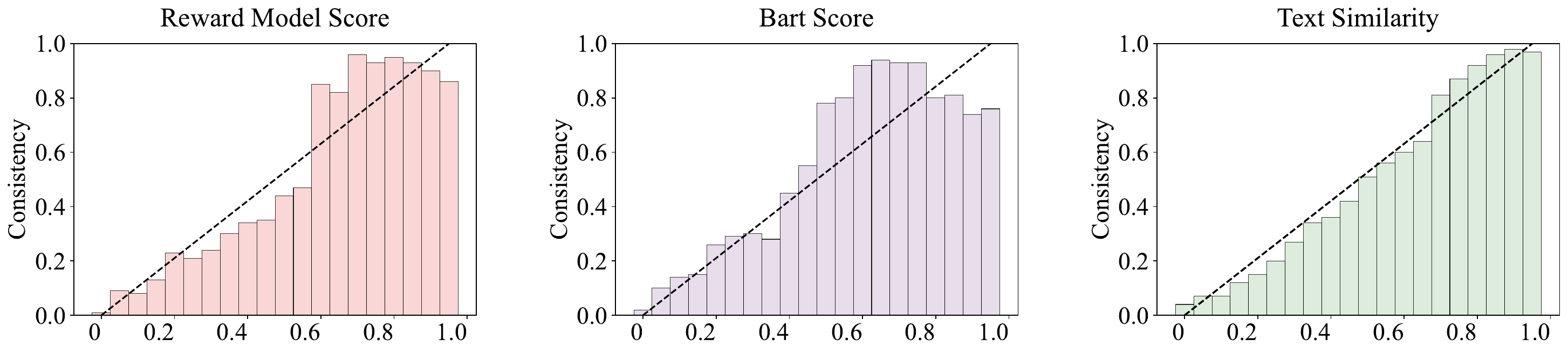}
    \caption{Comparison of consistency prediction signals.The left and middle plots show the relationship between score differences (LLM vs. DLM) from a reward model (Qwen2.5-PRM-7B) and BartScore, respectively, and human-annotated consistency labels. The right plot shows the same analysis using a reranker model (Qwen3-reranker-4B). The reranker score exhibits a stronger linear correlation with human labels, suggesting it better reflects true semantic consistency between responses.}
    \label{fig:consistency}
\end{figure*}

This instruction explicitly asks the model to consider the agreement between its own answer and that of a stronger LLM, nudging the internal representation to encode consistency-relevant information rather than only generic semantics.

We feed $x'$ into the DLM and take the final-layer hidden state of the EOS token as the semantic representation (Line~\ref{line:encode} in Algorithm~\ref{alg:consistency-routing}):
\begin{equation}
h_T := \text{DLM}(x')[\text{EOS}],
\label{eq:eos-pooling}
\end{equation}
where $\text{DLM}(x')$ denotes the sequence of hidden states at the last transformer layer and $\text{[EOS]}$ indexes the hidden state corresponding to the EOS token. This EOS-pooling scheme follows the design of recent instruction-based LLM embedding models, in which a short natural-language instruction and a special pooling token are used to obtain task-aligned sentence representations.

The resulting vector $h_T$ is then consumed by the router to predict the DLM--LLM consistency score and decide whether the query should be answered by the DLM or forwarded to the LLM. Importantly, the computation of $h_T$ is fully integrated into the DLM's prefilling stage: if the router decides to keep the query on the DLM, the key--value cache produced when encoding $x'$ can be directly reused for decoding. This eliminates redundant forward passes, avoids additional encoders (e.g., BERT-base) or external embedding APIs, and thus reduces both memory footprint and on-device latency.

\subsection{Lightweight Consistency Predictor (RQ2)}
\label{Lightweight Consistency Predictor}

\subsubsection{Motivation}

\begin{figure}[!t]
    \centering
    \includegraphics[width=0.99\columnwidth]{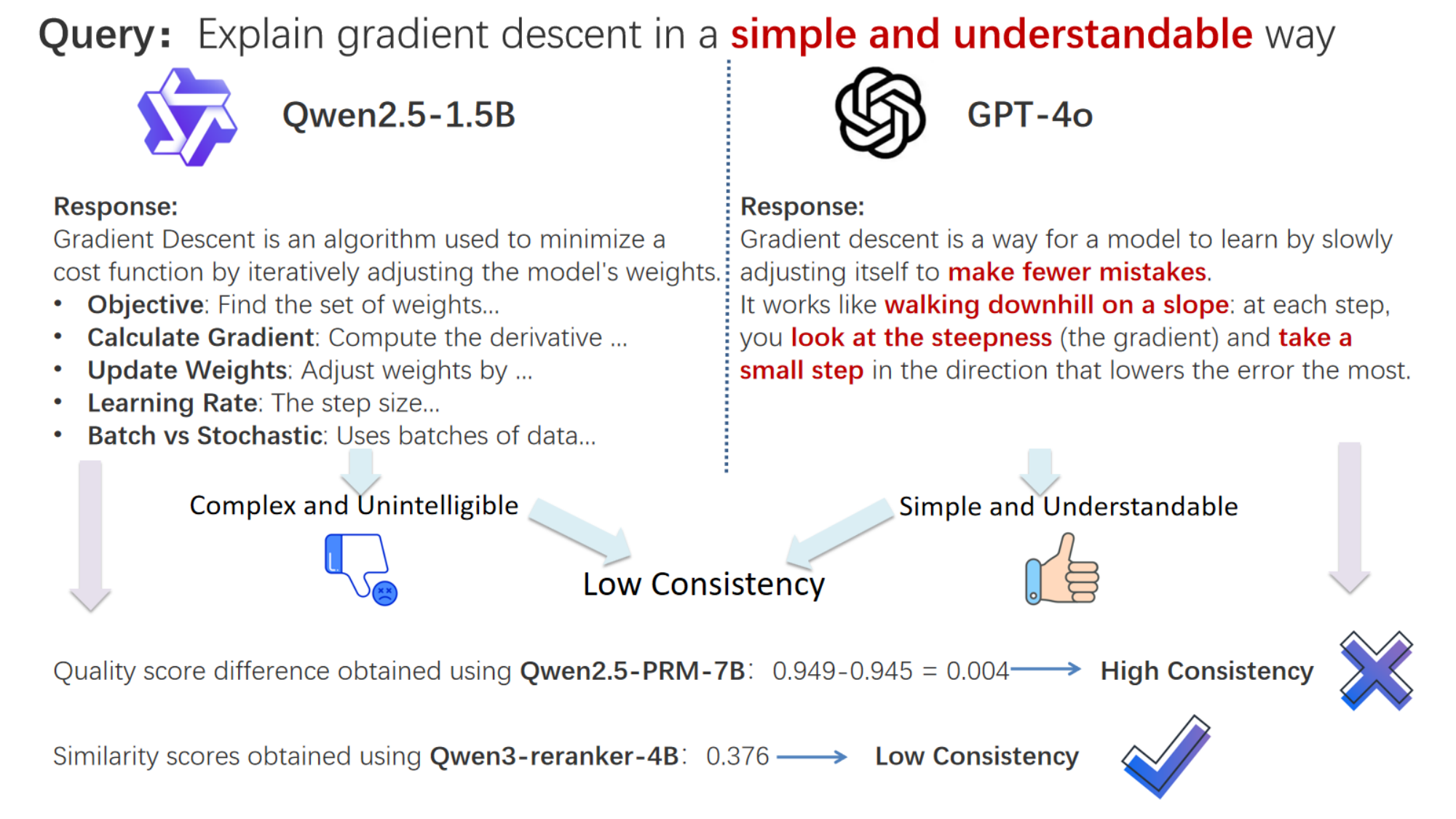} 
    \caption{Example illustrating the quality gap fails to reveal the semantic inconsistency. The reward model gives similar scores to both responses, while the reranker identifies their semantic mismatch.}
    \label{fig:consistency example}
\end{figure}

Existing routing methods usually rely on reward models or scalar scores that measure the gap between generated content and ground-truth answers, which may fail to capture semantic differences between responses. Conceptually, these approaches first map rich, high-dimensional textual semantics into a one-dimensional quality or reward score for each model, and then compute the difference of these scalar scores as the supervision signal. This two-step compression discards much of the fine-grained relational information between model outputs: both the DLM and CLM may receive similarly high scores while still conveying different factual content, so their score difference remains small even when their responses are semantically inconsistent. As a result, quality-gap–based supervision can systematically overlook semantic inconsistencies and lead to suboptimal routing decisions. As shown in Figure~\ref{fig:consistency example}, we use a representative example to illustrate this problem.

The example in Figure~\ref{fig:consistency example} shows that even if two responses differ significantly in terms of compliance with instructions and content consistency, the reward model may still assign similar high scores. This motivates us to shift from quality-based signals to semantic consistency. We evaluate how closely the DLM output aligns with that of a stronger model, thus directly supervising the routing signal based on meaning preservation.

To support multi-tier routing, we construct consistency supervision signals for both DLM--CLM and DLM--ELM response pairs. Since the construction procedures are identical for both cases, we describe the details using the DLM--CLM setting as a representative.

\subsubsection{Semantic Similarity as Consistency Supervision}

To quantify semantic consistency, we adopt a pre-trained reranker model $f_{\text{rank}}$, specifically, the Qwen3-Reranker-4B~\cite{qwen3embedding}. While rerankers are typically used to score and reorder a list of retrieved candidates based on their relevance to a given query~\cite{xiao2024c}, we repurpose it here to measure the semantic agreement between responses generated by different models. Given a query $x$ and two responses $y_s$ and $y_l$ from the DLM and CLM respectively, the reranker outputs a similarity score:
\begin{equation}
\mathrm{SIM}(x) := f_{\text{rank}}(y_s, y_l).
\end{equation}
This score captures whether the DLM response preserves the core semantics of the CLM response and serves as the primary supervision signal for training a consistency predictor.

Empirical analysis validates the effectiveness of semantic similarity as a training signal. As shown in Figure~\ref{fig:consistency}, the reranker-based score exhibits significantly higher agreement with human-labeled consistency judgments than traditional metrics like BartScore~\cite{hybridllm} or reward models~\cite{zooter}. This indicates that semantic similarity is a more reliable target for consistency-aware routing.

\subsubsection{Data Augmentation with Additional Signals}
While reranker-based similarity scores provide strong signals for semantic alignment, they may not fully capture all aspects of response consistency, particularly in ambiguous or under-specified cases. Moreover, relying solely on a single scoring model may introduce bias or reduce label diversity. To improve robustness and coverage, we incorporate complementary supervision signals via data augmentation:

\begin{itemize}
    \item \textbf{With references:} When the ground-truth $y^{*}$ is available for a query $x$, we apply rule-based supervision. If the DLM response $y_s$ is incorrect while the CLM response $y_l$ is correct (based on answer matching), we assign a hard consistency label $\mathrm{AUG}(x) = 0$; otherwise, $\mathrm{AUG}(x) = 1$ (including cases where both responses are correct, or both are incorrect, or DLM is correct while CLM is not).

    \item \textbf{Without references:} For queries without ground-truth, we introduce an advanced LLM as a judge to estimate the semantic agreement between the two responses. Specifically, we use a powerful open-ended LLM (DeepSeek V3) $f_{\text{judge}}$, providing it with the query $x$, the two responses $y_s$ and $y_l$, and a prompt asking whether the two responses convey the same meaning. The model returns a soft agreement score:
    \begin{equation}
        \mathrm{AUG}(x) := f_{\text{judge}}(x, y_s, y_l) \in [0,1].
    \end{equation}
\end{itemize}

These are combined into a unified consistency labels:
\begin{equation}
    S(x) := \alpha \cdot \mathrm{SIM}(x) + (1 - \alpha) \cdot \mathrm{AUG}(x),
\end{equation}
where a tunable weight $\alpha \in [0,1]$ controls the balance between general semantic similarity and augmented judgment.

\subsubsection{Predictor Training}

To support three-tier routing, we construct fused labels combining DLM--ELM and DLM--CLM consistency:
\begin{equation}
S_{\text{fused}}(x) := \beta \cdot S_{\text{cloud}}(x) + (1 - \beta) \cdot S_{\text{edge}}(x),
\end{equation}
where $S_{\text{cloud}}(x)$ and $S_{\text{edge}}(x)$ denote the consistency labels for DLM--CLM and DLM--ELM pairs respectively, and $\beta \in [0,1]$ controls their relative importance.

We use $h_T$ extracted from DLM as input to a lightweight MLP head $f_\theta$, producing the predicted consistency score:

\begin{equation}
\hat{S}(x) := f_\theta(h_T) \in [0, 1].
\end{equation}

The model is trained to minimize the mean squared error (MSE) loss against the fused soft label:

\begin{equation}
\mathcal{L}_{\text{mse}} = \| \hat{S}(x_i) - S_{\text{fused}}(x_i) \|^2.
\end{equation}

\subsection{Adaptive Threshold via Bayesian Optimization (RQ3)}
\label{Adaptive Threshold via Bayesian Optimization}

\paragraph{Motivation}

Due to the heterogeneity of queries, fixed routing thresholds $\tau_1$ and $\tau_2$ may not generalize well. In practice, query complexity, sensitivity, or task type vary greatly. Some queries, such as simple open-ended questions, can be handled relatively tolerantly by DLM or ELM even with low consistency scores; while others, such as mathematical problems, are more risky and require the highest response quality even if the prediction consistency appears high. This difference suggests that a one-size-fits-all threshold strategy may lead to suboptimal routing decisions in heterogeneous environments.

To address this challenge, we adopt an adaptive threshold selection mechanism based on Bayesian optimization~\cite{mockus2005bayesian}. Instead of using a single global pair of thresholds, we assign distinct routing thresholds $(\tau_1^{(k)}, \tau_2^{(k)})$ for each category of queries, allowing the system to better reflect query-specific preferences and risk profiles. Algorithm~\ref{alg:adaptive-thresholds} summarizes this cluster-based threshold learning procedure and its online adaptation.

\begin{algorithm*}[htbp]
\caption{Cluster-Based Threshold Optimization with Online Bayesian Adaptation}
\label{alg:adaptive-thresholds}
\begin{algorithmic}[1]
\REQUIRE Historical queries $\{x_i\}$ with representations $\{h_i\}$, utility $\mathcal{U}_k(\tau_1,\tau_2)$, offline BO budget $T_{\text{off}}$, online update interval $M$
\ENSURE Centroids $\{c_k\}_{k=1}^K$, thresholds $\{(\tau_1^{(k)}, \tau_2^{(k)})\}_{k=1}^K$

\STATE \textbf{// Offline clustering and threshold learning}\label{line:offline-begin}
\STATE Determine $K$ via the elbow method and run K-means on $\{h_i\}$ to obtain clusters $\{\mathcal{C}_k\}_{k=1}^K$ and centroids $\{c_k\}_{k=1}^K$
\FOR{$k = 1,\dots,K$}
    \STATE Initialize $\mathcal{D}_k \leftarrow \emptyset$
    \FOR{$t = 1,\dots,T_{\text{off}}$}
        \STATE Fit GP surrogate $f_k$ on $\mathcal{D}_k$
        \STATE Select $(\tau_1,\tau_2)$ by maximizing EI under $\tau_1 > \tau_2$, $\tau_1,\tau_2 \in [0,1]$
        \STATE Evaluate $u \leftarrow \mathcal{U}_k(\tau_1,\tau_2)$ on $\mathcal{C}_k$ and update $\mathcal{D}_k \leftarrow \mathcal{D}_k \cup \{((\tau_1,\tau_2),u)\}$
    \ENDFOR
    \STATE $(\tau_1^{(k)},\tau_2^{(k)}) \leftarrow \arg\max_{((\tau_1,\tau_2),u)\in\mathcal{D}_k} u$
\ENDFOR\label{line:offline-end}

\vspace{0.3em}
\STATE \textbf{// Online adaptation (streaming phase)}\label{line:online-begin}
\STATE Initialize global counter $t \leftarrow 0$
\FOR{each incoming query $x$}
    \STATE $t \leftarrow t + 1$
    \STATE Obtain $h$ and predicted score $\hat{S}(x)$ using Algorithm~\ref{alg:consistency-routing}, Lines~\ref{line:construct-input}--\ref{line:assign-cluster}
    \STATE Assign cluster $k^* \leftarrow \arg\min_k \|h - c_k\|_2$ and route $x$ with $(\tau_1^{(k^*)},\tau_2^{(k^*)})$
    \STATE After observing correctness, latency, and cost, compute per-query utility $u(x)$
    \STATE Update $\mathcal{D}_{k^*} \leftarrow \mathcal{D}_{k^*} \cup \{((\tau_1^{(k^*)},\tau_2^{(k^*)}),u(x))\}$
    \IF{$t \bmod M = 0$}
        \FOR{$k = 1,\dots,K$}
            \STATE Fit GP surrogate $f_k$ on $\mathcal{D}_k$
            \STATE Run a small number of BO steps on cluster $\mathcal{C}_k$ using $f_k$ and $\mathcal{D}_k$ to refresh $(\tau_1^{(k)},\tau_2^{(k)})$
        \ENDFOR
    \ENDIF
\ENDFOR\label{line:online-end}
\end{algorithmic}
\end{algorithm*}

\paragraph{Cluster-Based Bayesian Optimization}
To identify categories of queries with different routing preferences, we treat the queries used for training the router as historical queries, and cluster them based on their semantic representations $h_T$ using classical K-means. The number of clusters $K$ is automatically determined using the elbow method~\cite{kaufman2009elbow}. For each cluster $\mathcal{C}_k$ with centroid $c_k$, we learn optimal thresholds $(\tau_1^{(k)}, \tau_2^{(k)})$ ($\tau_1^{(k)} > \tau_2^{(k)}$ and $\tau_1^{(k)}, \tau_2^{(k)} \in [0,1]$) that maximizes system utility:
\begin{equation}
\mathcal{U}_k(\tau_1^{(k)}, \tau_2^{(k)}) := \lambda_1 \cdot \text{Acc} - \lambda_2 \cdot \text{Latency} - \lambda_3 \cdot \text{Cost},
\end{equation}
where $\text{Acc}$ denotes the average response correctness under the current threshold-based routing, $\text{Latency}$ represents the average inference delay, and $\text{Cost}$ reflects the estimated computation cost. The weights $\lambda_1$, $\lambda_2$, and $\lambda_3$ control the trade-off among these objectives.

For each query $x$ in cluster $\mathcal{C}_k$, the predicted consistency score $\hat{S}(x)$ determines the selected model (DLM, ELM, or CLM) based on the corresponding thresholds $(\tau_1^{(k)}, \tau_2^{(k)})$. The accuracy term $\text{Acc}$ is computed as the average correctness of selected model responses within the cluster. If a ground-truth reference answer is available, a response is considered correct if it exactly matches the reference; otherwise, we use an advanced LLM judge to assess the semantic appropriateness of the response. The $\text{Latency}$ term includes end-to-end inference time, covering both model execution and communication delays. The $\text{Cost}$ term is estimated based on the number of activated parameters and the length of generated tokens, following the compute cost model in~\cite{wilkins2024cost}. Notably, the ratios $\lambda_1 / \lambda_2$ and $\lambda_1 / \lambda_3$ ($\lambda_1, \lambda_2, \lambda_3 > 0$) reflect the system’s preference between accuracy and efficiency: a larger ratio favors quality, while a smaller ratio emphasizes speed and cost reduction.

We perform Gaussian Process-based Bayesian optimization~\cite{mockus2005bayesian} to automatically determine the optimal thresholds $(\tau_1^{(k)}, \tau_2^{(k)})$ for each cluster $\mathcal{C}_k$. At each iteration, we fit a surrogate model to the observed utility values $\mathcal{U}_k$ evaluated at previous threshold pairs, and use an acquisition function (Expected Improvement~\cite{jones1998efficient}) to propose the next threshold configuration to evaluate. This iterative process continues until the evaluation budget is exhausted, and is summarized in the offline phase of Algorithm~\ref{alg:adaptive-thresholds}, Lines~\ref{line:offline-begin}--\ref{line:offline-end}.

\paragraph{Online Adaptation of Cluster Thresholds}
The above procedure learns cluster-specific thresholds $(\tau_1^{(k)}, \tau_2^{(k)})$ from historical data. In practice, however, traffic patterns, query difficulty, and user preferences may drift over time. To keep the thresholds aligned with the current environment, we further endow our framework with an online adaptation mechanism based on incremental Bayesian optimization.

For each cluster $\mathcal{C}_k$, we maintain a Gaussian Process surrogate $f_k$ over the threshold space and a set of observed utility samples $\mathcal{D}_k = \{((\tau_1, \tau_2), u)\}$. The offline optimization phase initializes $\mathcal{D}_k$ with utility values $\mathcal{U}_k(\tau_1^{(k)}, \tau_2^{(k)})$ evaluated on historical queries. At deployment time, when a new query $x$ arrives, we first assign it to a semantic cluster $k^*$ and route it using the current thresholds $(\tau_1^{(k^*)}, \tau_2^{(k^*)})$. After the response is generated, we observe its correctness, latency, and cost, and compute a per-query utility
\begin{equation}
u(x) = \lambda_1 \cdot \mathbb{I}[\text{correct}] - \lambda_2 \cdot \text{Latency}(x) - \lambda_3 \cdot \text{Cost}(x),
\end{equation}
where $\mathbb{I}[\text{correct}]$ is the indicator function that equals $1$ if the response is judged correct and $0$ otherwise. In our experiments, correctness is determined using the same criterion as in the offline phase: 
exact match against the ground-truth answer when available, and an advanced LLM judge otherwise.In a real deployment, this term can be instantiated with any task-specific binary success signal (e.g., automatic verifiers or delayed user feedback), without changing the optimization procedure. This quantity can be viewed as a stochastic sample of the cluster-level utility $\mathcal{U}_{k^*}$.  We then append the pair $((\tau_1^{(k^*)}, \tau_2^{(k^*)}), u(x))$ to $\mathcal{D}_{k^*}$ and incrementally update the surrogate $f_{k^*}$.

Periodically (after accumulating a fixed number of new queries in $\mathcal{D}_k$), we run Gaussian Process-based Bayesian optimization on each cluster in the background: using $f_k$ and the acquisition function Expected Improvement, (EI)~\cite{jones1998efficient}, we propose new candidate threshold pairs and update $(\tau_1^{(k)}, \tau_2^{(k)})$ if they yield higher estimated utility. Here EI measures the expected amount by which a candidate $(\tau_1,\tau_2)$ can improve over the best utility observed so far. This incremental procedure interleaves exploitation of the current best thresholds with exploration of promising alternatives, and allows the routing thresholds to continuously adapt to evolving query distributions without adding latency to the per-query routing path. The overall online adaptation loop corresponds to the streaming phase of Algorithm~\ref{alg:adaptive-thresholds}, Lines~\ref{line:online-begin}--\ref{line:online-end}.

\paragraph{Inference-Time Threshold Assignment}
At runtime, once we obtain the semantic representation $h_T$ of the input query $x$, we determine its routing thresholds by assigning it to the nearest semantic cluster. We measure the Euclidean distance between $h_T$ and each cluster centroid $c_k$, and select the closest cluster (Line~\ref{line:assign-cluster} in Algorithm~\ref{alg:consistency-routing}):
\begin{equation}
k^* := \arg\min_k \|h_T - c_k\|_2.
\end{equation}
We then apply the optimized thresholds $(\tau_1^{(k^*)}, \tau_2^{(k^*)})$ associated with cluster $\mathcal{C}_{k^*}$ for routing decision (Line~\ref{line:retrieve-thresholds} in Algorithm~\ref{alg:consistency-routing} and the routing step in Algorithm~\ref{alg:adaptive-thresholds}, Lines~\ref{line:online-begin}--\ref{line:online-end}). This enables context-sensitive thresholds to adjust to the semantic characteristics of the query, resulting in more precise trade-offs between performance, latency, and cost.

\section{Experiments}

\subsection{Experimental Settings}

\paragraph{Data Construction}
We conduct experiments on the RouterBench dataset ~\cite{hu2024routerbench}, which consists of 36.5K queries from 8 NLP datasets in both Chinese and English. For each query, we take it as input and collect the outputs of small and large language models. To construct training labels for the consistency predictor, we combine two sources: (1) Qwen3-Reranker-4B~\cite{qwen3embedding} for semantic similarity between DLM and higher-tier outputs, and (2) data augmentation via correctness-based rules (with references) or DeepSeek V3~\cite{liu2024deepseek} as a judge (without references).  For queries with known correct answers, we label cases where the DLM is incorrect but the CLM is correct as inconsistent. For queries without known correct answers, we apply an advanced model to estimate response consistency. The final label is obtained by averaging the similarity score and the augmentation score.

\paragraph{Datasets}
Our main results focus on MMLU~\cite{hendrycks2020mmlu} (general knowledge), GSM8K~\cite{cobbe2021GSM8K} (math reasoning), HumanEval~\cite{chen2021humaneval} (code generation) and MT-Bench~\cite{mtbench} (conversation, judged by GPT-4o).

\paragraph{Baseline Algorithms}
We compare our approach with several representative baselines. The LLM-only baseline routes all queries to the large language model regardless of difficulty or cost. The DLM-only baseline handles all queries with the small language model, maximizing efficiency but potentially sacrificing accuracy. The Edge-only baseline processes all queries using an edge-optimized model with a moderate resource footprint. We further compare against learning-based routing methods. RouteLLM (BERT)~\cite{ong2024routellm} uses a fine-tuned BERT encoder to obtain a representation of each query, and feeds the [CLS] token into a logistic regression classifier to predict the probability that the DLM performs better than the CLM. RouteLLM (SW ranking) embeds the query with a text embedding model, retrieves its nearest historical query in the embedding space, and follows the routing decision associated with that most similar past query. MixLLM~\cite{wang2025mixllm} encodes the query using a tag-enhanced BERT model and feeds the representation into multiple predictors (e.g., for quality, latency, and cost); a policy then decides whether to route to the DLM or CLM based on the estimated latency–utility trade-off.

\paragraph{Deployment Environment and Model Configuration}
To simulate a realistic hierarchical deployment scenario, we use different hardware platforms and language model configurations to represent device-side, edge-side, and cloud-side environments. On the device, we deploy Qwen3-1.7B~\cite{yang2025qwen3} as the device language model (DLM), running on an Intel Core i5-12500H CPU paired with an NVIDIA RTX 3050 GPU, which mimics the capabilities of a typical consumer-grade endpoint. On the edge, we use a server equipped with a single NVIDIA RTX A6000 GPU to host the edge language model (ELM), specifically Qwen3-14B, representing a high-capability on-premise or near-edge setup. In the cloud, we deploy Qwen3-32B across two RTX A6000 GPUs to simulate access to large-scale language models with near-unlimited compute resources. This three-tier configuration reflects a practical deployment scenario with ascending model capacity and latency from device to cloud, and is used consistently throughout our experiments.

\paragraph{Network Configuration for Edge and Cloud Links}
\label{sec:network-config}

Based on our previous measurements of wide-area mobile networks and edge networks~\cite{edge_imc21}, we simulated end-to-edge and end-to-cloud connections by configuring the device-side downlink/uplink bandwidth, packet loss rate, one-way latency, and DNS latency. We simulated two network environments: "good" and "bad". 

In the "good" network condition, the device-to-edge link was configured with: 10,000 Kbps downlink bandwidth, 5,000 Kbps uplink bandwidth, 0.1\% bidirectional packet loss rate, one-way latency of 40 ms (downlink) and 20 ms (uplink), and a DNS latency of 50 ms; the device-to-cloud link was configured with: 8,000 Kbps downlink bandwidth, 4,000 Kbps uplink bandwidth, 0.1\% packet loss rate, one-way latency of 80 ms and 40 ms, and a DNS latency of 70 ms. This represents a relatively stable and well-configured connection. In the "bad" network condition, we simulated connections with reduced bandwidth, higher packet loss rates, and higher latency. The device-to-edge link was configured with: 2,000 Kbps downlink bandwidth, 500 Kbps uplink bandwidth, 1\% packet loss rate, one-way latency of 120 ms and 80 ms, and a DNS latency of 200 ms. The device-to-cloud link had even stricter limitations: 800 Kbps downlink bandwidth, 200 Kbps uplink bandwidth, 3\% packet loss rate, one-way latency of 250 ms and 200 ms, and a DNS latency of 400 ms. 

In the "Network Condition Sensitivity" experiment, we used three settings: "good", "bad" and "bad"→"good", which allowed us to observe how ConsRoute adjusts its routing thresholds as the underlying network improves. All other experiments were conducted under "good" network conditions.

\begin{figure*}[!t]
\centering
\subfloat[GSM8K]{
    \includegraphics[width=0.23\textwidth]{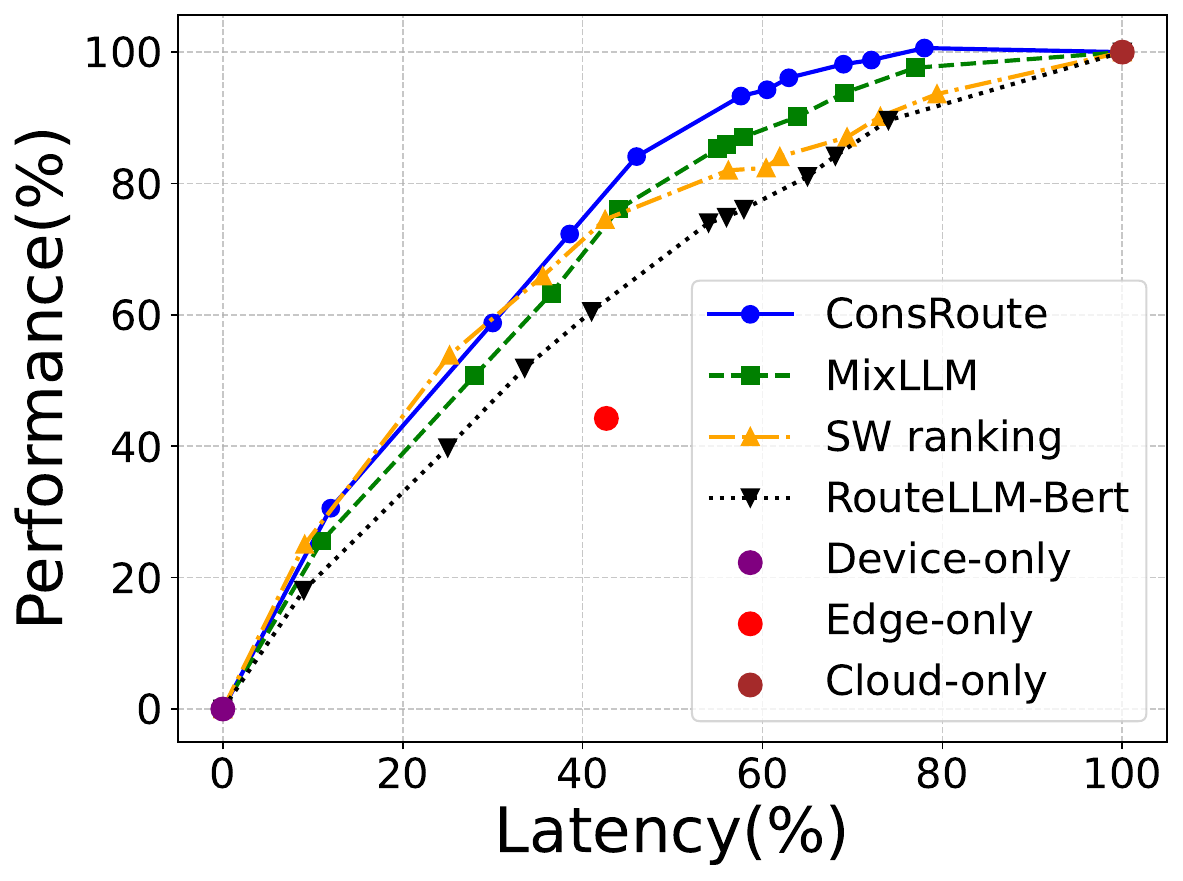}%
    \label{fig:gsm8k-latency-main}
}
\hfil
\subfloat[MMLU]{
    \includegraphics[width=0.23\textwidth]{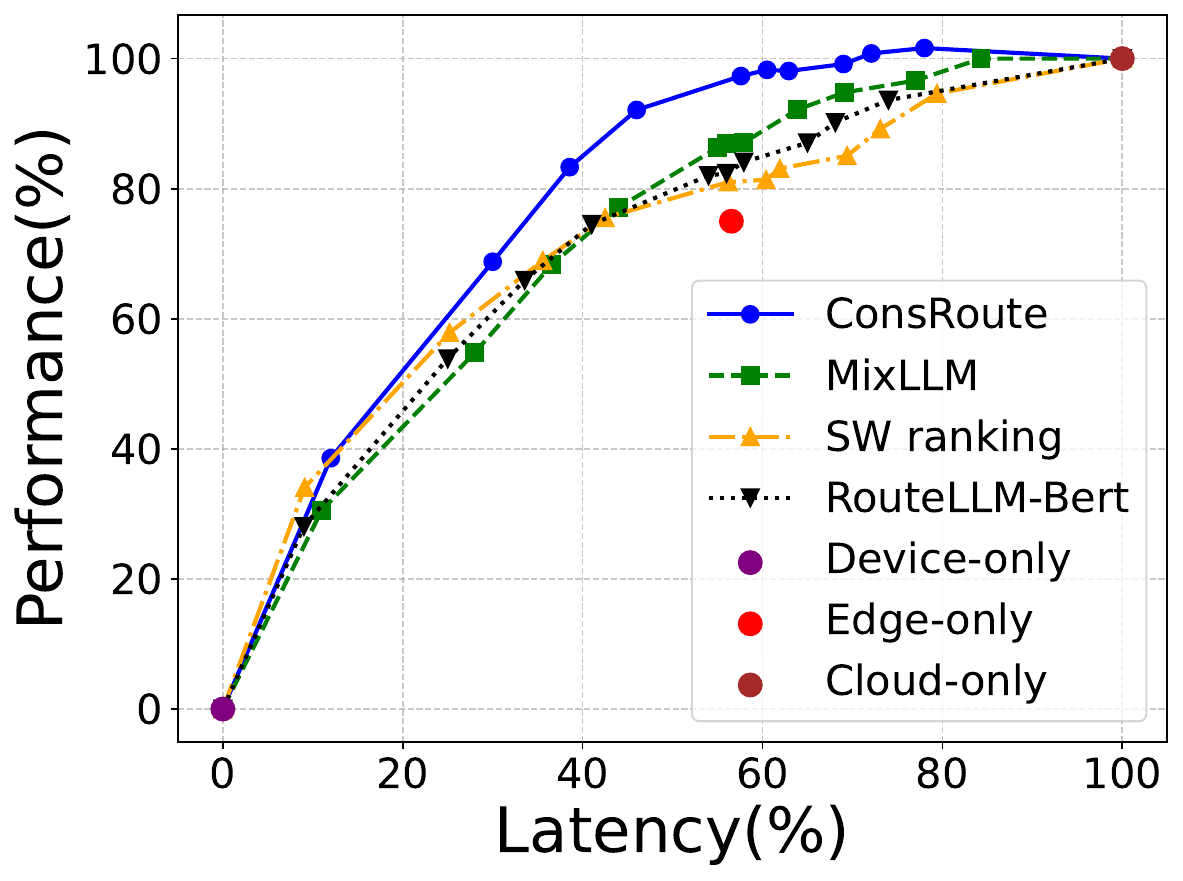}%
    \label{fig:mmlu-latency-main}
}
\hfil
\subfloat[HumanEval]{
    \includegraphics[width=0.23\textwidth]{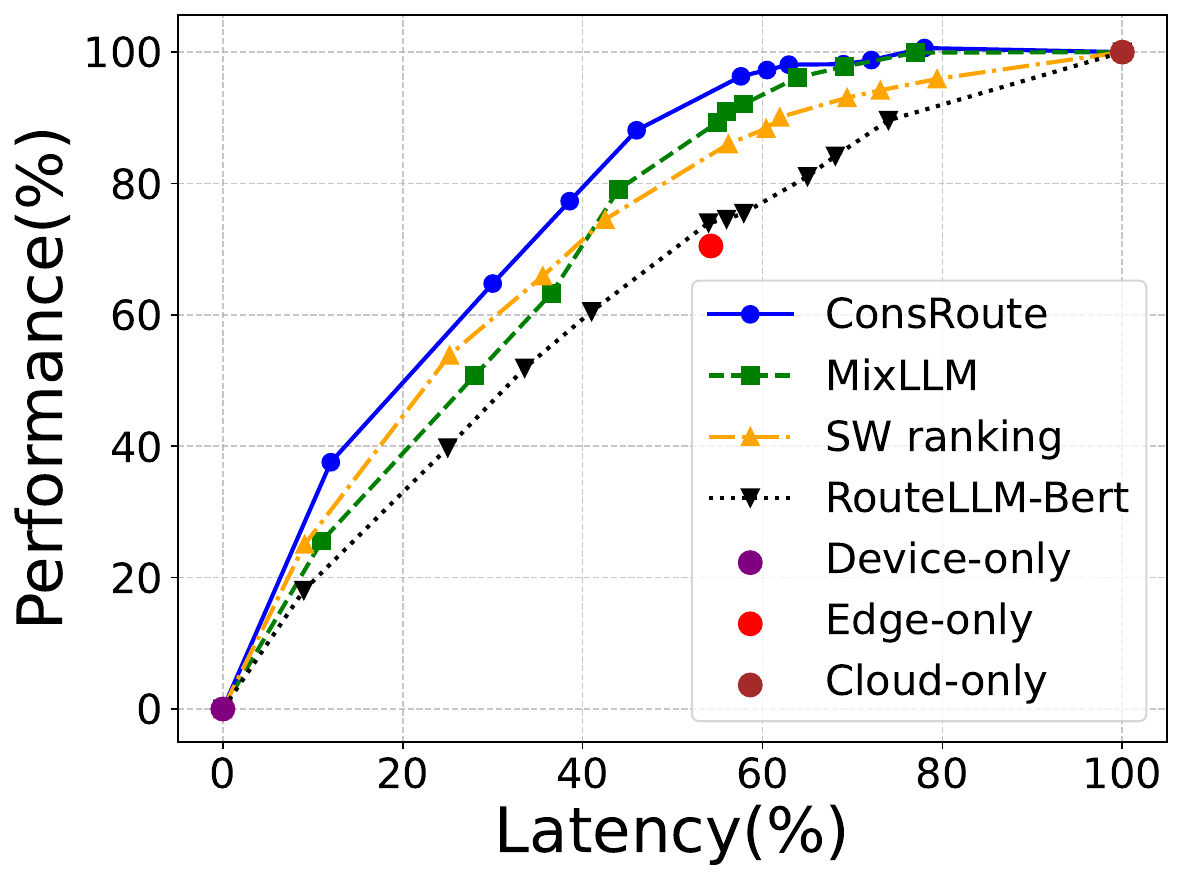}%
    \label{fig:humaneval-latency-main}
}
\hfil
\subfloat[MT-Bench]{
    \includegraphics[width=0.23\textwidth]{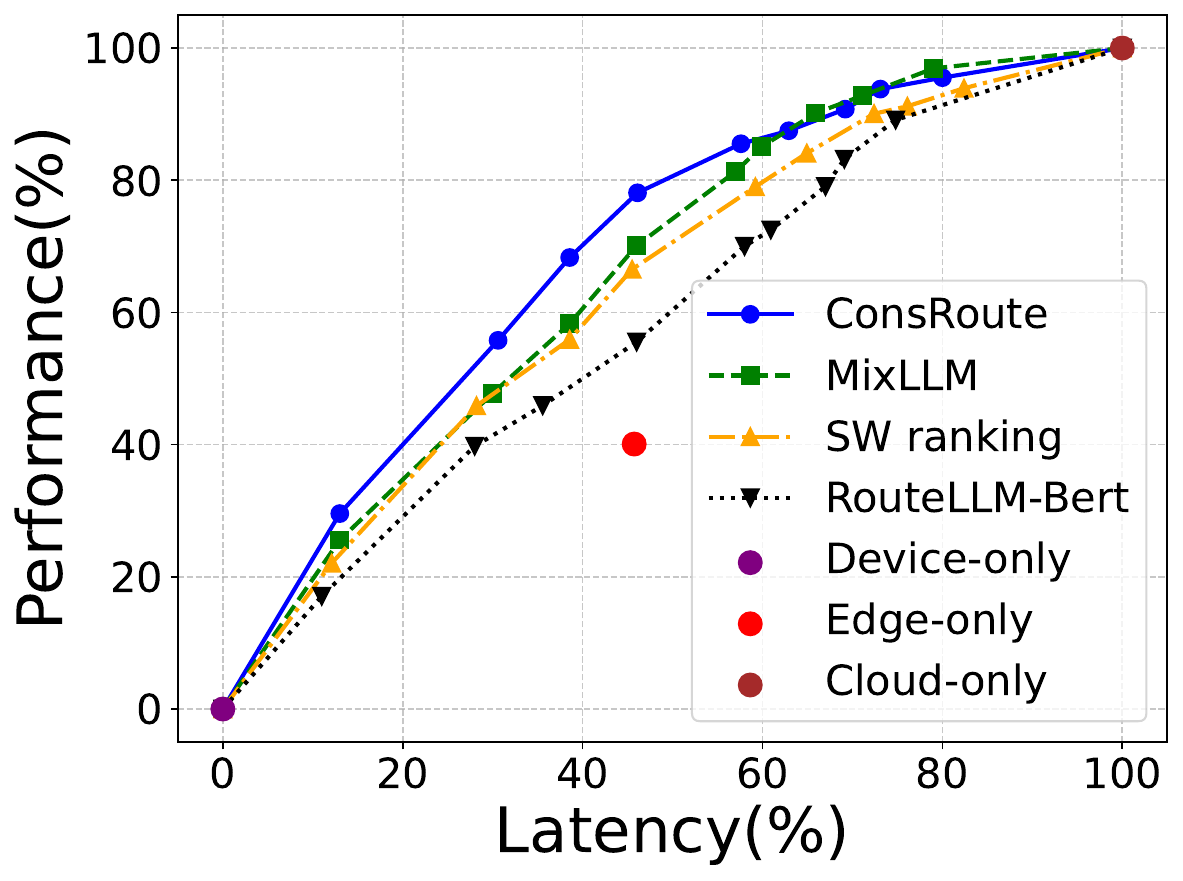}%
    \label{fig:mtbench-latency-main}
}

\subfloat[GSM8K]{
    \includegraphics[width=0.23\textwidth]{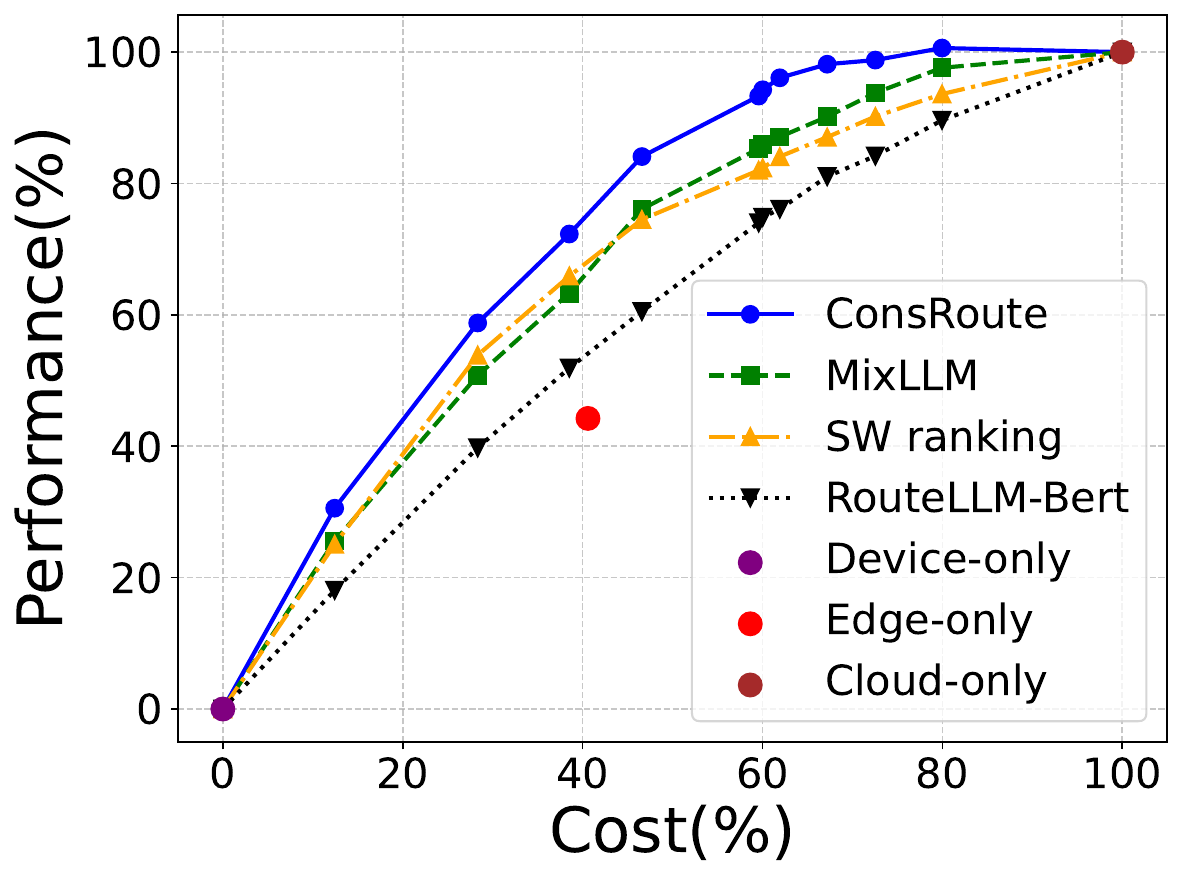}%
    \label{fig:gsm8k-cost-main}
}
\hfil
\subfloat[MMLU]{
    \includegraphics[width=0.23\textwidth]{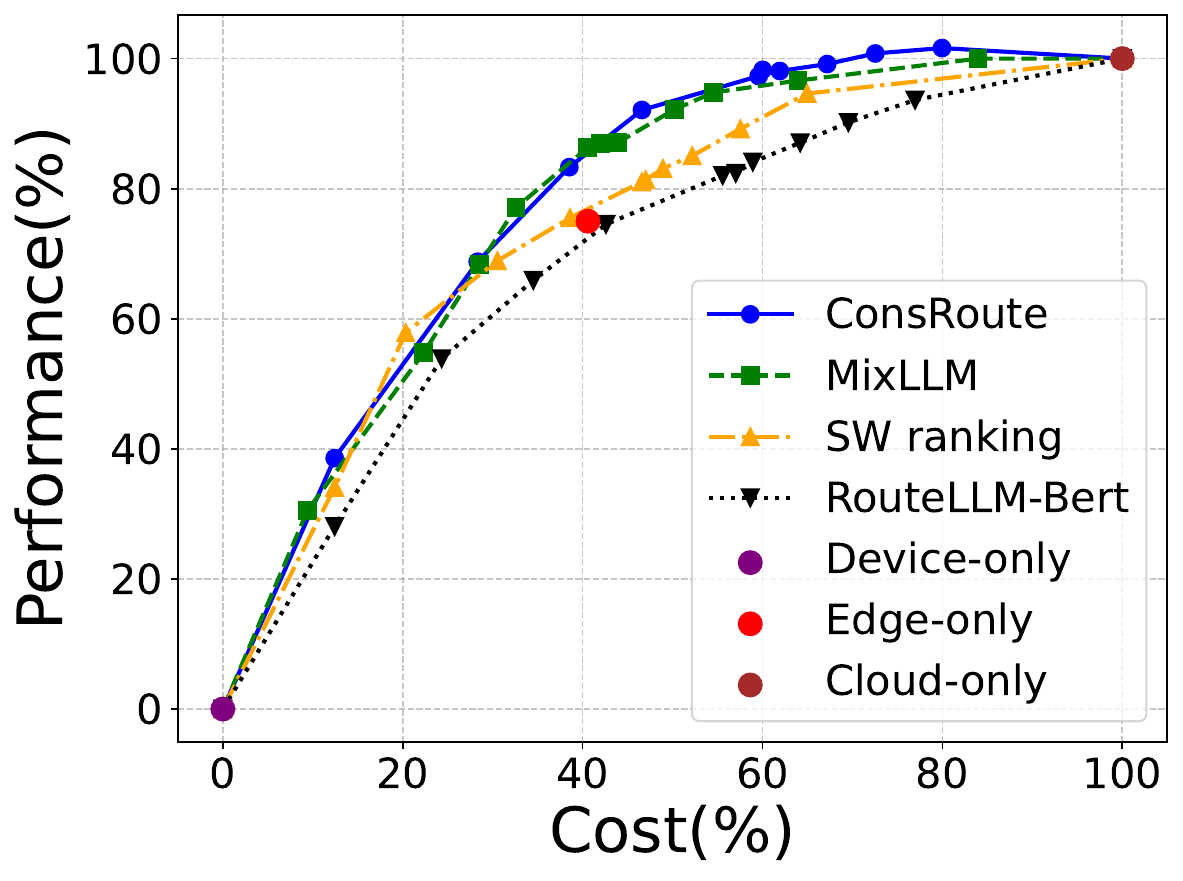}%
    \label{fig:mmlu-cost-main}
}
\hfil
\subfloat[HumanEval]{
    \includegraphics[width=0.23\textwidth]{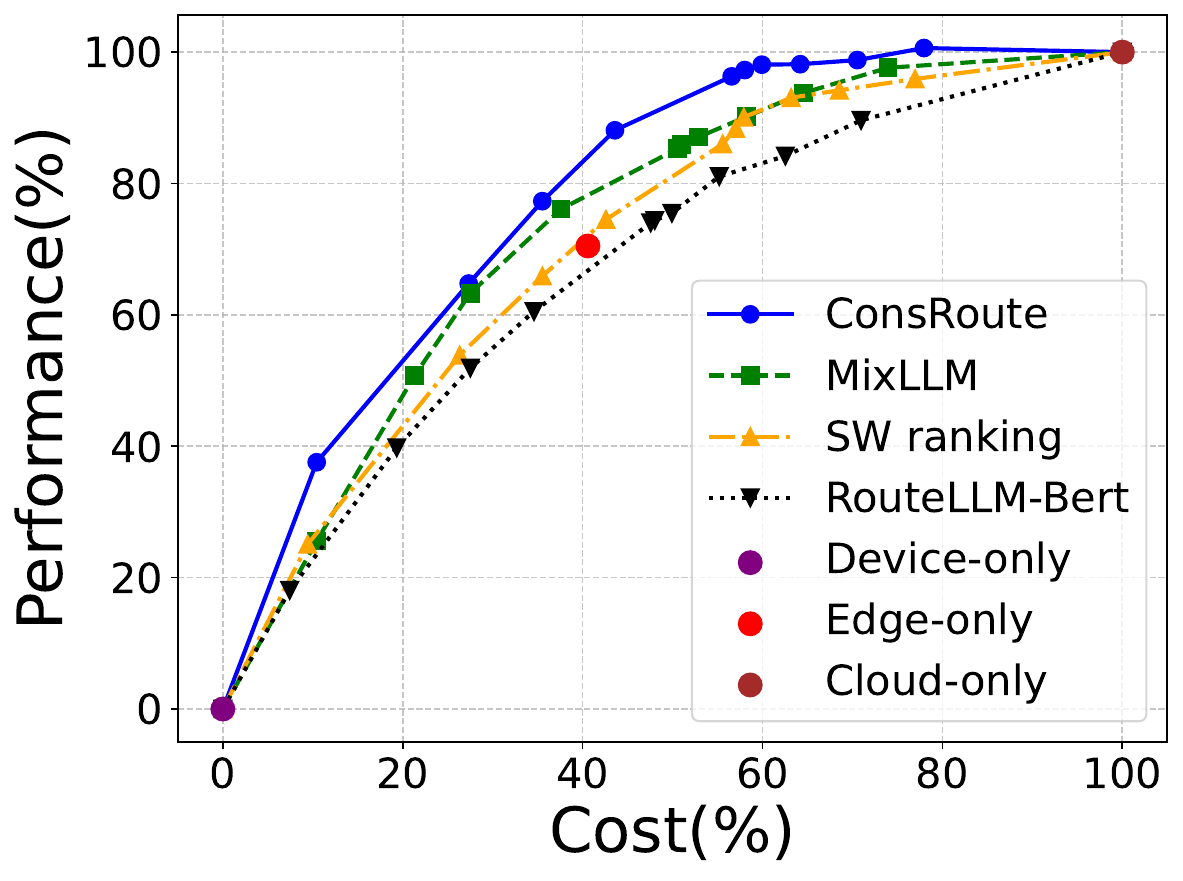}%
    \label{fig:humaneval-cost-main}
}
\hfil
\subfloat[MT-Bench]{
    \includegraphics[width=0.23\textwidth]{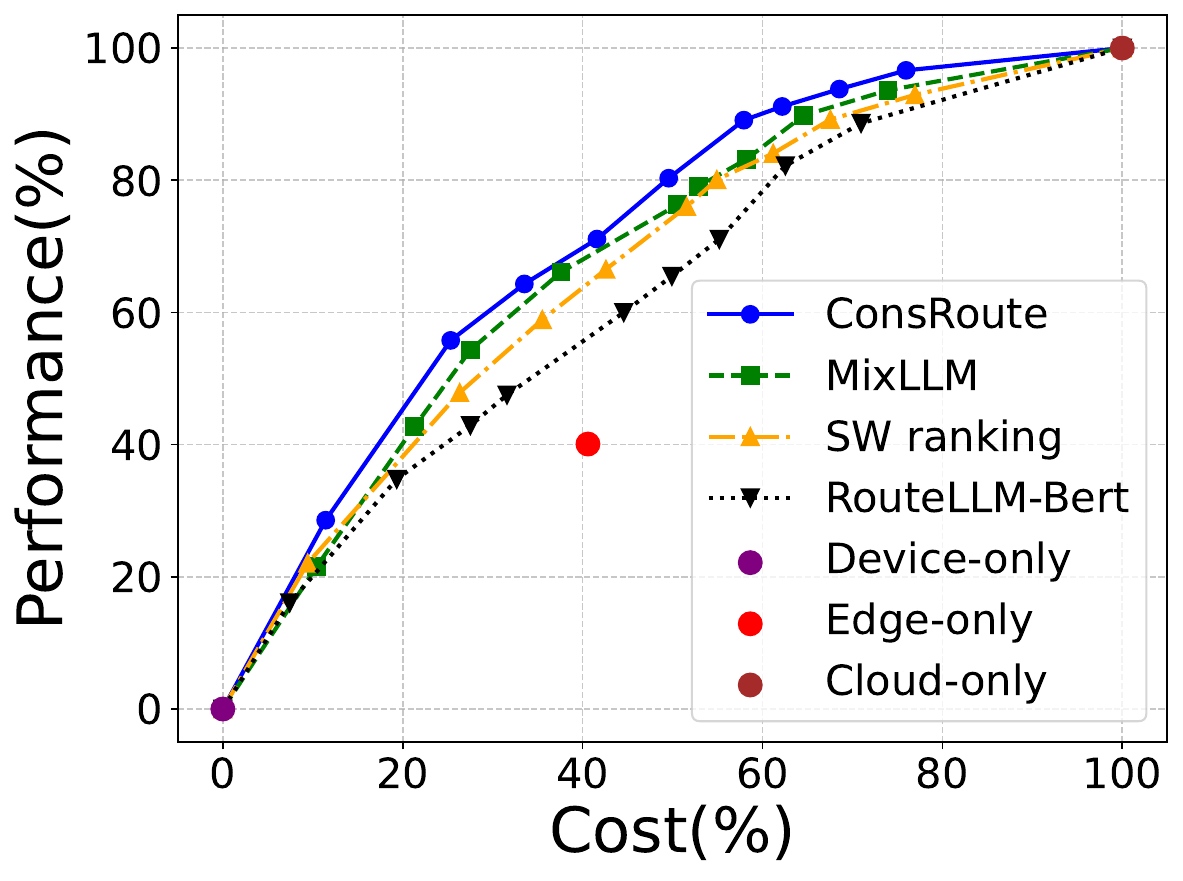}%
    \label{fig:mtbench-cost-main}
}

\caption{Comparison of accuracy, latency, and cost across benchmarks. ConsRoute achieves near-cloud accuracy with significantly reduced latency and cost.}
\label{fig:main-result}
\end{figure*}

\subsection{Overall Performance and Cross-Model Adaptability}

\subsubsection{Overall Results}

As shown in Figure~\ref{fig:gsm8k-latency-main} to Figure~\ref{fig:humaneval-latency-main}, and Figure~\ref{fig:gsm8k-cost-main} to Figure~\ref{fig:humaneval-cost-main}, ConsRoute consistently outperforms baseline methods in the latency-performance trade-off across all three benchmarks—GSM8K, MMLU, HumanEval and MT-Bench—and achieves competitive or superior results in the cost-performance trade-off in most cases.

Under the current experimental setup, ConsRoute achieves near-cloud performance ($\geq$95\%) using only 60\%-65\% of the latency required by a cloud-only strategy. In comparison, baseline methods such as MixLLM and RouteLLM typically require 70\%-85\% of the cloud latency to reach similar accuracy levels. In terms of cost, ConsRoute achieves comparable performance using only about 70\% of the cloud-only cost, and outperforms most baselines across a wide range of routing thresholds.

To enable fair and interpretable comparison, latency, cost, and score are all normalized relative to three fixed reference points: (0, 0) for device-only (DLM), (100, 100) for cloud-only (CLM), and a mid-point representing edge-only (ELM) based on actual measured latency and cost. Each method generates a set of routing decisions under varying latency or cost constraints, which are then plotted as a curve reflecting its trade-off behavior.

These results demonstrate that ConsRoute is able to make better use of hierarchical model capacities in cloud-edge-device collaborative inference, reaching high-quality predictions more efficiently than existing approaches.

\subsubsection{Cross-Family Cloud--Edge--Device Deployments}

To verify that our framework does not rely on homogeneity, we further 
consider a heterogeneous deployment in which each tier is served by a 
different model family: LLaMA-3.2-3B on the device, Qwen3-14B on the edge,  and DeepSeek-V3~\cite{liu2024deepseek} in the cloud (accessed via API). For DeepSeek-V3, a mixture-of-experts model, we estimate inference cost using the number of activated parameters (37B) rather than the full model size (671B). This setting mimics realistic multi-vendor deployments where models differ in architecture, tokenizer, and training pipeline.

\begin{figure}[!t]
\centering
\subfloat[End-to-end latency vs.\ accuracy.]{
    \includegraphics[width=0.85\linewidth]{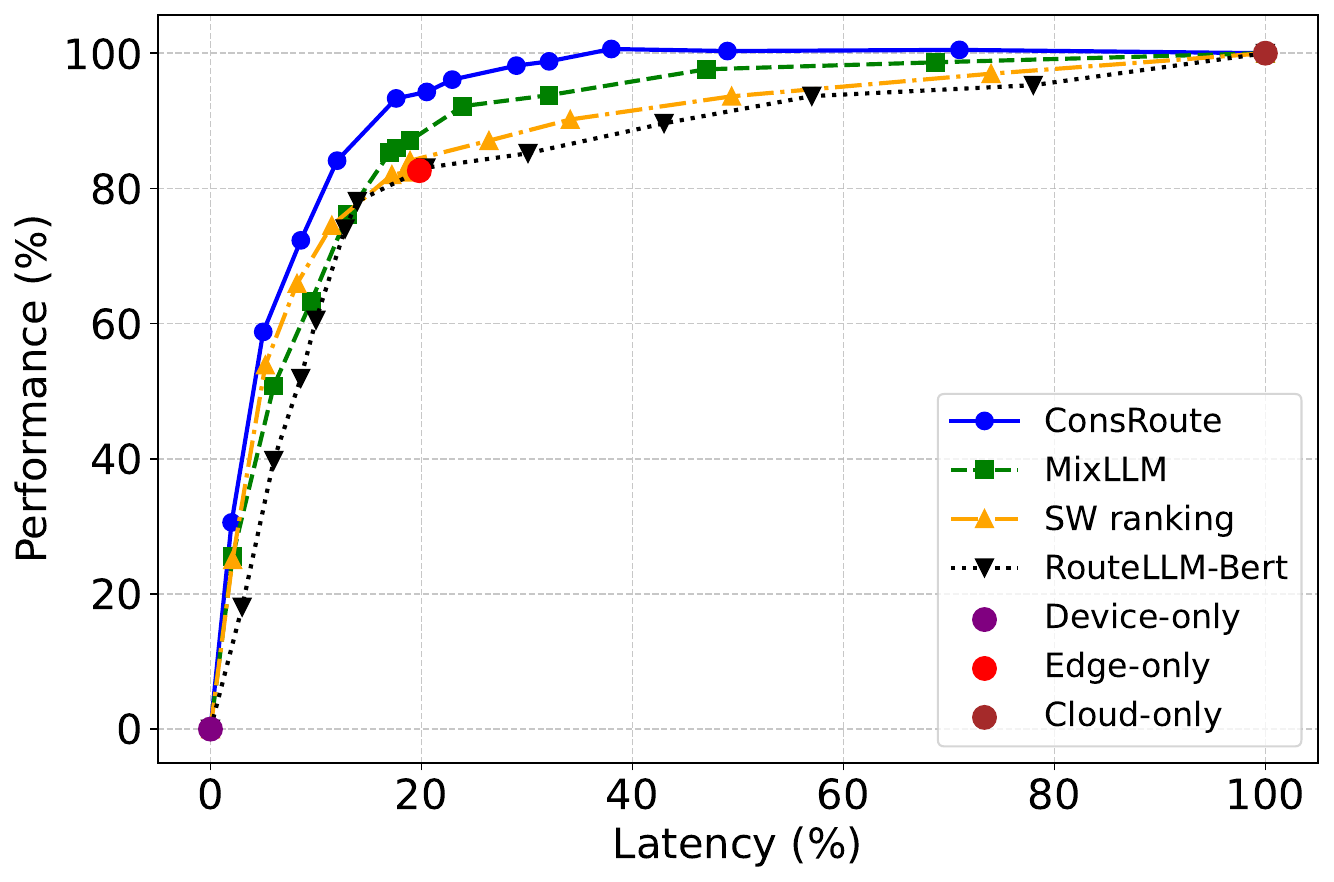}%
    \label{fig:label-latency}
}
\hfil
\subfloat[Inference cost vs.\ accuracy.]{
    \includegraphics[width=0.85\linewidth]{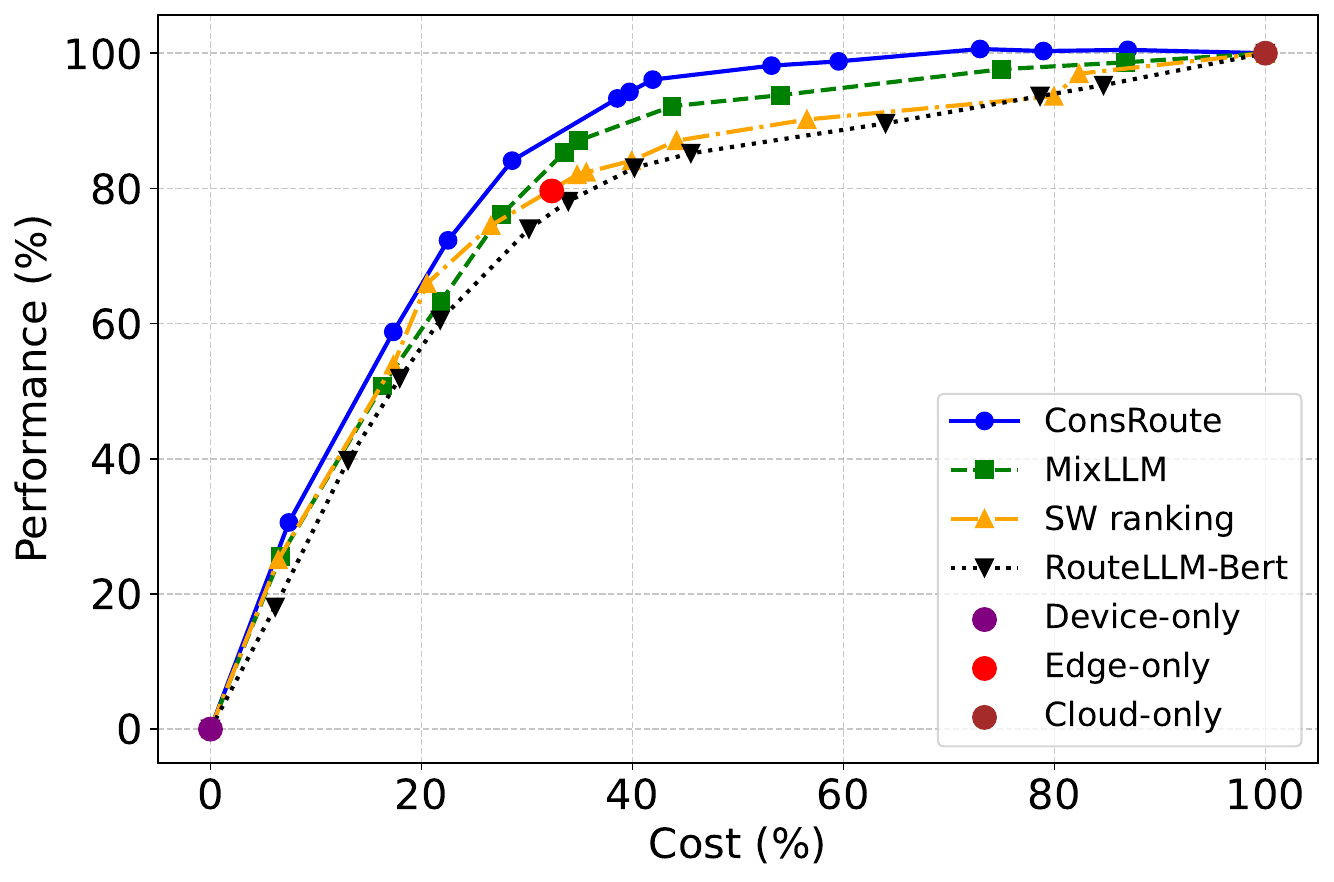}%
    \label{fig:label-cost}
}
\caption{Adaptability of ConsRoute under heterogeneous model deployments on GSM8K. }
\label{fig:adaptable}
\end{figure}

Results on GSM8K (Figure~\ref{fig:adaptable}) show that ConsRoute still achieves favorable accuracy--latency and accuracy--cost trade-offs under this cross-family configuration, comparable to those observed with the homogeneous stack. This demonstrates that our routing framework can generalize across heterogeneous model combinations and remains effective even when the cloud, edge, and device tiers are served by different LLM families.

\subsection{Analysis of Routing Signals and DLM Representations (RQ1 \& RQ2)}

\subsubsection{Routing Latency and Device-Side Extra Resource Overhead}

To assess deployment feasibility, we compare our routing latency and device-side overhead with RouteLLM and MixLLM. For RouteLLM-SW Ranking, latency includes OpenAI embedding API calls and local vector retrieval. RouteLLM-BERT and MixLLM incur latency from BERT encoding and MLP classification, with RouteLLM-BERT further requiring on-device execution of a 110M-parameter BERT-base model—introducing notable memory and compute costs on resource-limited devices.

\begin{figure}[!t]
    \centering
    \includegraphics[width=0.85\columnwidth]{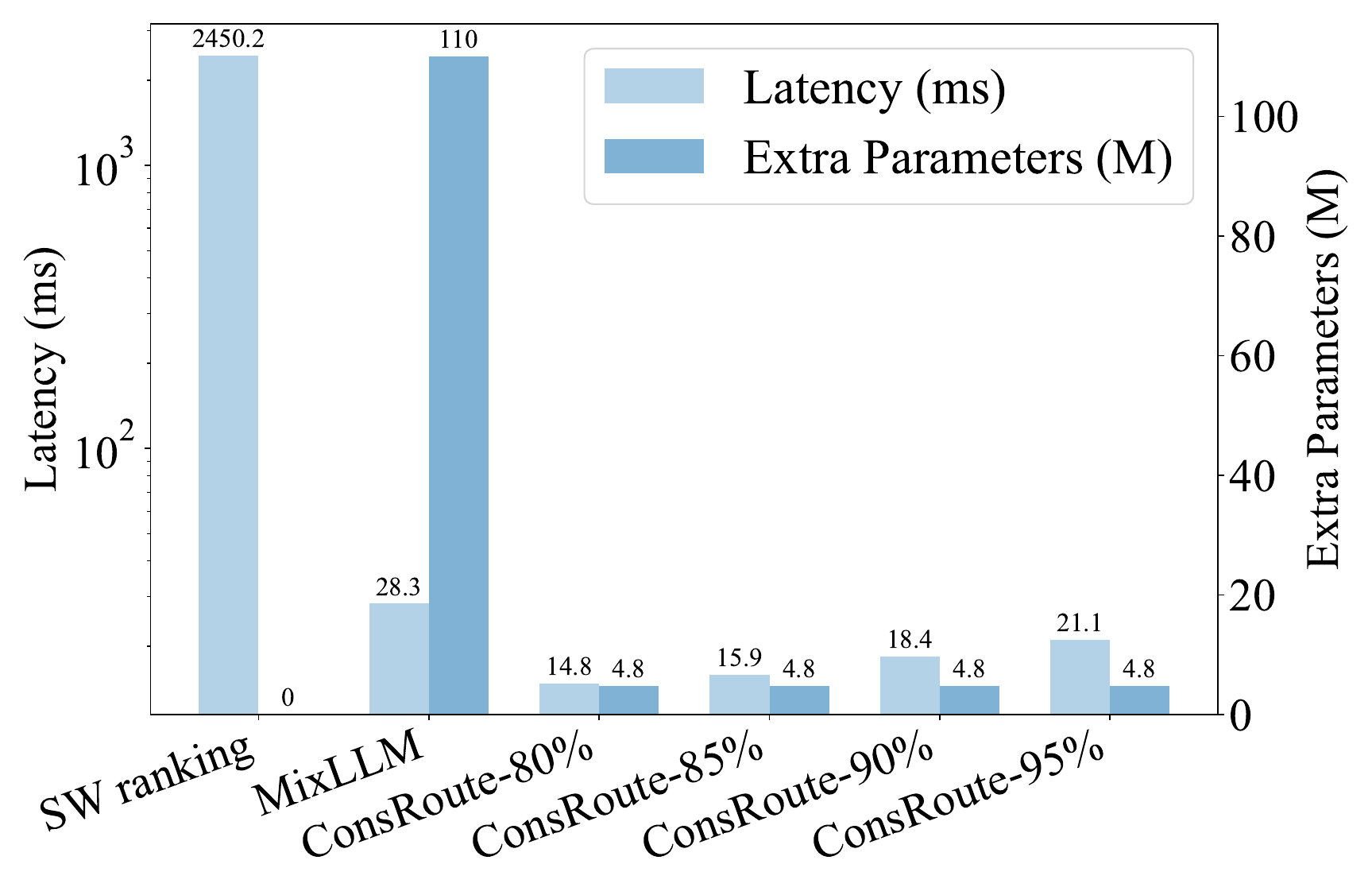} 
    \caption{Routing Latency and additional device-side resource overhead of different routers. ``ConsRoute-\textit{x}\%'' denotes the latency and overhead when the system achieves \textit{x}\% of the CLM-only accuracy. RouteLLM and MixLLM have the same latency and additional overhead with different accuracy rates.}
    \label{fig:router latency}
\end{figure}

In contrast, our method significantly reduces routing latency and resource overhead. When routing to a higher-tier model, latency includes only the DLM prefill (for hidden state extraction) and the lightweight MLP prediction. If the DLM is selected for generation, its prefill is part of inference, and routing delay reduces to just the MLP's inference time. The MLP itself has only ~5M parameters, adding negligible computation and memory overhead. This compact design ensures minimal impact on edge devices.

Figure~\ref{fig:router latency} illustrates the routing latency and additional device-side resource overhead (measured by extra parameter count) for different methods. Our approach consistently achieves faster routing speed compared to baselines, while introducing negligible additional overhead on the device, making it well-suited for real-time applications.

\subsubsection{Effect of Prompting Strategies on Routing}
We further conduct a study to examine how different prompting strategies applied to the DLM affect routing performance, contrasting explicit routing decisions made by the DLM with implicit representations fed to a separate router. As shown in Table~\ref{tab:table2}, on the GSM8K dataset we compare four variants: (1) explicit selection without CoT, where the DLM is directly asked to choose the execution tier (device, edge, or cloud) and outputs a discrete decision without any reasoning; (2) explicit selection with CoT, where the DLM first explains its routing decision via chain-of-thought and then outputs the final choice; (3) implicit selection without prompt, where we simply feed the last-token hidden state of the DLM without any additional prompt into the router; and (4) implicit selection with prompt (our method), where a fixed consistency-oriented instruction is appended and the last token hidden state is used as the router input. The exact prompts used for the explicit routing baselines (with and without CoT) are provided in Appendix A.

The results demonstrate that prompt-guided implicit representation is the most effective and efficient design. The naive implicit variant without prompt achieves the lowest accuracy , indicating that generic next-token hidden states are not sufficiently aligned with the consistency prediction task. Adding an explicit prompt and using last token hidden state substantially improves accuracy, while keeping router latency almost unchanged, confirming that our prompt primarily reshapes the internal representation without introducing extra routing overhead. The non-CoT variant attains moderate accuracywith relatively high latency , whereas enabling chain-of-thought slightly improves accuracy but incurs prohibitive routing latency. These findings validate our design choice of using prompt-guided implicit representations: it captures consistency-aware semantics better than unprompted hidden states and achieves a more favorable accuracy--latency trade-off than explicit DLM-based routing.

\begin{table}[tbp]
    \caption{Effect of prompting strategies on routing performance}
    \label{tab:table2}
    \centering
    {\renewcommand{\arraystretch}{1.5}%
    \setlength{\tabcolsep}{6pt}%
    \begin{tabular}{lcc}
        \hline
        Method & Accuracy  & Router Latency (ms) \\
        \hline
        Explicit selection (w/o CoT)   & 85.79 &  36.2  \\
        Explicit selection (w/ CoT)    & 86.44 & 1868.2 \\
        Implicit selection (no prompt) & 82.34 &  20.4  \\
        Implicit selection (prompted)  & 87.92 &  20.9  \\
        \hline
    \end{tabular}%
    }
\end{table}

\subsubsection{Comparison of Supervision Sources}

To investigate the impact of different training signals for consistency-aware routing, 
we compare three types of supervision labels: reward model scores (from Qwen2.5-PRM-7B), BartScore~\cite{hybridllm}, and reranker-based (from Qwen3-Reranker-4B) semantic similarity.For each label type, we train the same routing architecture using identical data and evaluate their performance under consistent conditions.

\begin{figure}[!t]
\centering
\subfloat[Latency vs Accuracy]{
    \includegraphics[width=0.47\linewidth]{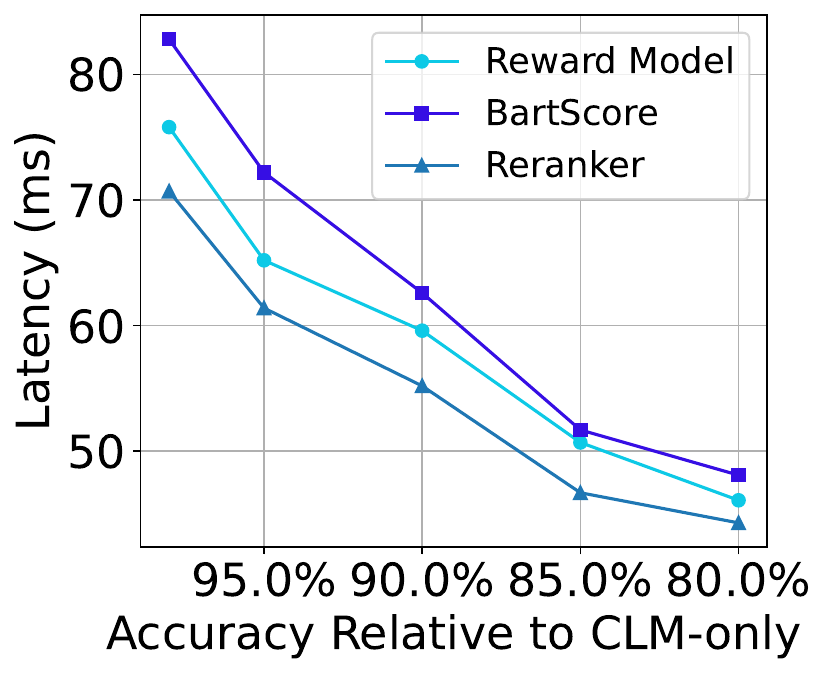}%
    \label{fig:label-latency}
}
\hfil
\subfloat[Cost vs Accuracy]{
    \includegraphics[width=0.47\linewidth]{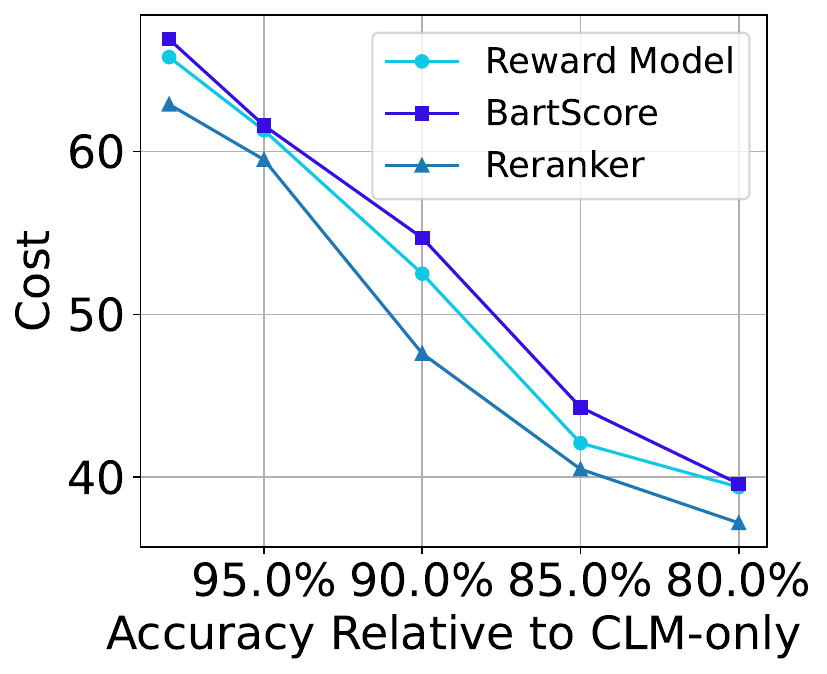}%
    \label{fig:label-cost}
}
\caption{Impact of supervision source on routing efficiency. Each curve shows the latency or cost required to reach a target CLM-only accuracy.}
\label{fig:different-labels}
\end{figure}

As shown in Figure~\ref{fig:different-labels}, reranker-based supervision consistently achieves lower latency and cost at the same level of performance. These results confirm that semantic similarity is a more faithful and effective training signal for consistency-aware routing.

\subsection{Threshold Adaptation and System Trade-offs (RQ3)}

\subsubsection{Online Threshold Adaptation}

\begin{figure}[!t]
    \centering
    \includegraphics[width=0.95\columnwidth]{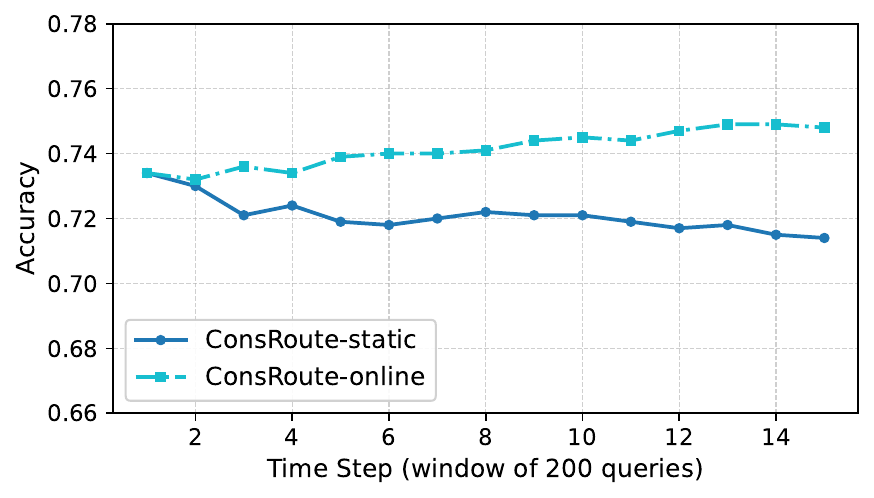}
    \caption{Online threshold adaptation of ConsRoute. ConsRoute-online maintains consistently higher accuracy as the query distribution drifts.}
    \label{fig:online-adapt}
\end{figure}

To evaluate the effectiveness of the proposed online adaptation mechanism, we simulate a streaming deployment scenario on a held-out subset of RouterBench that is not used for training or offline threshold tuning. Queries in this subset are ordered to form an input stream and divided into non-overlapping windows of 200 consecutive requests. Each window is treated as one time step on the horizontal axis in Figure~\ref{fig:online-adapt}, and we record the routing accuracy within each window as the performance at that time step. After processing a window, the online variant is allowed to update cluster-specific thresholds using the new feedback, whereas the static variant keeps thresholds fixed.

We compare two variants of our framework under this setting. ConsRoute-static uses the cluster-specific thresholds learned in the offline phase and keeps them unchanged throughout the entire stream. ConsRoute-online starts from the same offline thresholds but, after each window of 200 requests, performs an incremental Bayesian optimization step to refine thresholds for clusters that received new data. Both variants share the same consistency predictor, clustering assignments, and deployment configuration.

Figure~\ref{fig:online-adapt} plots the routing accuracy over time, where the horizontal axis denotes the index of the 200-request window and the vertical axis shows the average accuracy within that window. As the query stream progressed, the accuracy of ConsRoute-static decreased slightly, while ConsRoute-online, after several threshold updates, was able to recover and maintain higher accuracy. These results demonstrate that online threshold adaptation can effectively track distributional changes and improve long-term routing performance.

\subsubsection{Sensitivity to Network Conditions}

To study how ConsRoute reacts to varying network conditions in a cloud--edge--device setting, we simulate three types of network environments (\emph{Good}, \emph{Bad}, and \emph{Bad}$\rightarrow$\emph{Good}) on an online evaluation subset of RouterBench. Guided by prior measurement studies on mobile and edge networks~\cite{edge_imc21}, we emulate device-to-edge and device-to-cloud connections by configuring their downlink/uplink bandwidth, packet loss rate, one-way delay, and DNS delay. The \emph{Bad}$\rightarrow$\emph{Good} condition starts with the Bad configuration in the first half of the query stream and switches to the Good configuration in the second half, mimicking a recovery from a congested to a normal network. The detailed parameter settings for the edge and cloud links under the Good and Bad profiles are provided in Experimental Settings. All other settings are identical to those in the Online Threshold Adaptation experiment.

\begin{figure}[!t]
    \centering
    \includegraphics[width=0.95\columnwidth]{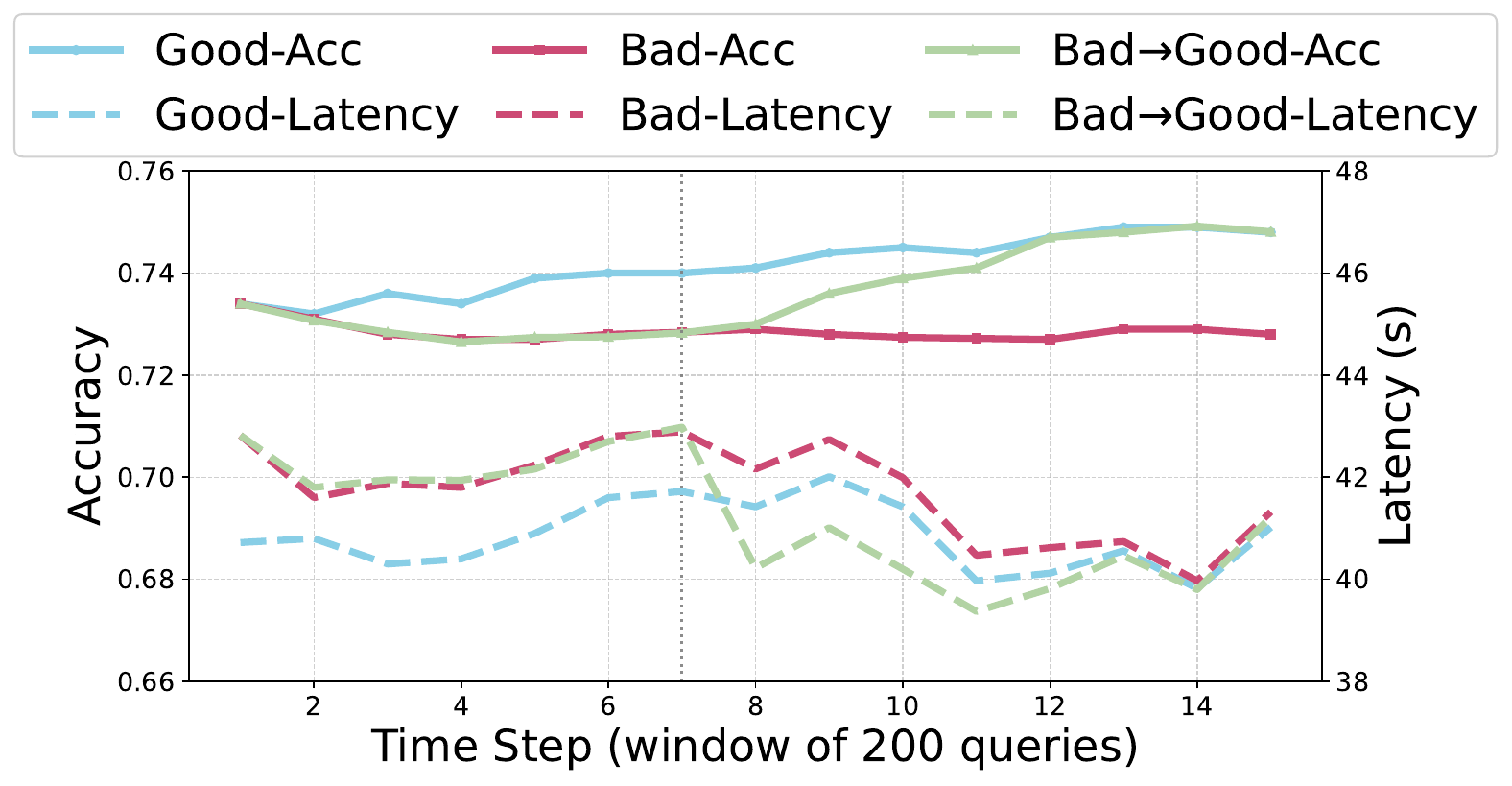}
    \caption{Network condition sensitivity of ConsRoute. Accuracy (left axis) and end-to-end latency (right axis) over time under three network profiles: \textit{Good}, \textit{Bad}, and \textit{Bad}$\rightarrow$\textit{Good}.  In the \textit{Bad}$\rightarrow$\textit{Good} setting, the network switches from the Bad profile to the Good profile at time step 7.}
    \label{fig:network-sensitivity}
\end{figure}

Figure~\ref{fig:network-sensitivity} shows six curves: three accuracy trajectories (Good-Acc, Bad-Acc, Bad$\rightarrow$Good-Acc) and three latency trajectories (Good-Latency, Bad-Latency, Bad$\rightarrow$Good-Latency) over time. Under the Good condition, ConsRoute-online balances quality and communication cost: it is willing to route a non-trivial fraction of queries to edge and cloud models, achieving higher accuracy. Under the Bad condition, the increased communication delay is heavily penalized in the utility function, and the incremental Bayesian optimization gradually shifts cluster-specific thresholds toward more conservative escalation. As a result, the fraction of queries served by the on-device DLM increases, leading to a noticeable reduction in average latency, accompanied by a modest drop in accuracy that remains within an acceptable range.

In the Bad$\rightarrow$Good condition, we observe a two-phase adaptive behavior. During the initial Bad phase, the trends resemble those of the pure Bad condition: thresholds are adjusted to favor on-device routing, decreasing latency while slightly degrading accuracy. Once the network switches back to the Good configuration, the new latency feedback is incorporated into the utility estimates, and Bayesian optimization begins to relax the thresholds, making the system more willing to route difficult queries to edge or cloud models. These results demonstrate that ConsRoute-online can automatically adapt its routing thresholds to changing network conditions, achieving intuitive accuracy--latency trade-offs without manual retuning or retraining of the router.

\subsubsection{Sensitivity to Utility Hyperparameters}

To understand how the choice of utility weights on correctness, latency, and cost in the Bayesian optimization utility function shapes the routing behavior, we conduct a sensitivity study on the HumanEval dataset. We normalize the accuracy weight $\lambda_1$ and vary the relative penalty on latency and cost by adjusting the ratios $\kappa_1 = \lambda_1 / \lambda_2$ and $\kappa_2 = \lambda_1 / \lambda_3$.  For each hyperparameter setting, we record the fraction of queries routed to device, edge, and cloud, as well as the resulting end-to-end accuracy.

\begin{figure}[!t]
    \centering
    \includegraphics[width=0.95\columnwidth]{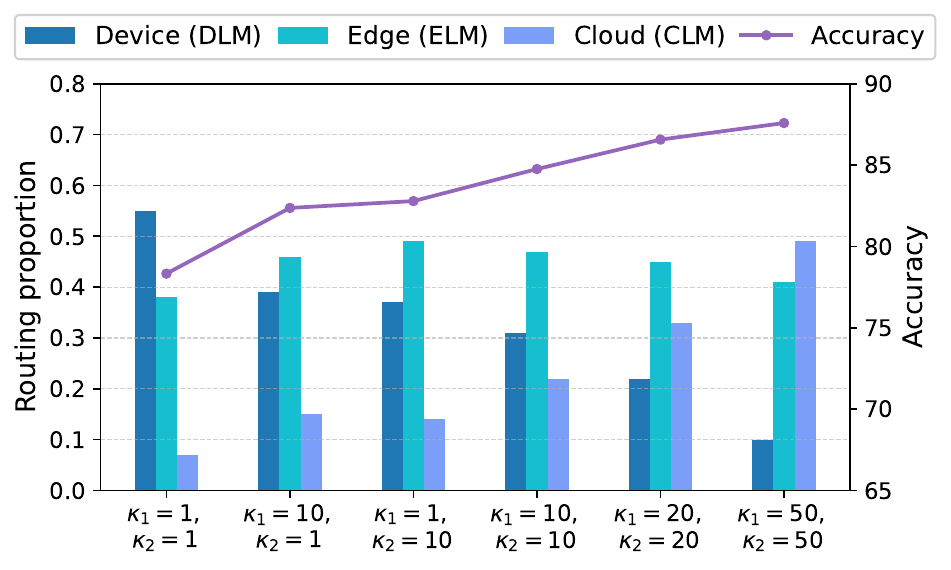}
    \caption{Effect of utility weights on routing behavior and accuracy on HumanEval. Bars show the proportions of queries routed to device, edge, and cloud under different weight settings, and the line reports the corresponding accuracy.  $\kappa_1$ and $\kappa_2$ denote the normalized weights $\kappa_1 = \lambda_1 / \lambda_2$ and $\kappa_2 = \lambda_1 / \lambda_3$.}
    \label{fig:utility-sensitivity}
\end{figure}

Figure~\ref{fig:utility-sensitivity} summarizes the results. The bar plots show the proportions of queries assigned to the DLM, ELM, and CLM under different utility configurations, while the overlaid line reports the corresponding accuracy. When the utility heavily penalizes latency and cost (small $\kappa_1$ and $\kappa_2$), ConsRoute routes the majority of queries to the DLM and ELM, achieving the lowest accuracy but the highest efficiency. As the weight on correctness increases, the routing gradually shifts from device and edge to the cloud tier, and the overall accuracy improves monotonically. In the most accuracy-oriented setting, a substantial fraction of queries are escalated to the CLM, yielding the highest accuracy at the expense of higher latency and cost. These results confirm that the proposed utility-based formulation provides a smooth and interpretable knob for trading off quality versus efficiency in the cloud-edge-device collaboration.

\subsection{Other Ablation Studies}

To assess the contributions of key components in our routing framework, we conduct additional ablation experiments on the HumanEval dataset. We study the effect of two design choices: label augmentation during training, and dynamic thresholding via Bayesian optimization. The comparison includes three configurations: (1) a minimal version without label augmentation and with a fixed global threshold; (2) a variant with label augmentation but a static threshold; and (3) our full model, which incorporates both.

\begin{figure}[!t]
\centering
\subfloat[Performance vs Latency]{
    \includegraphics[width=0.88\linewidth]{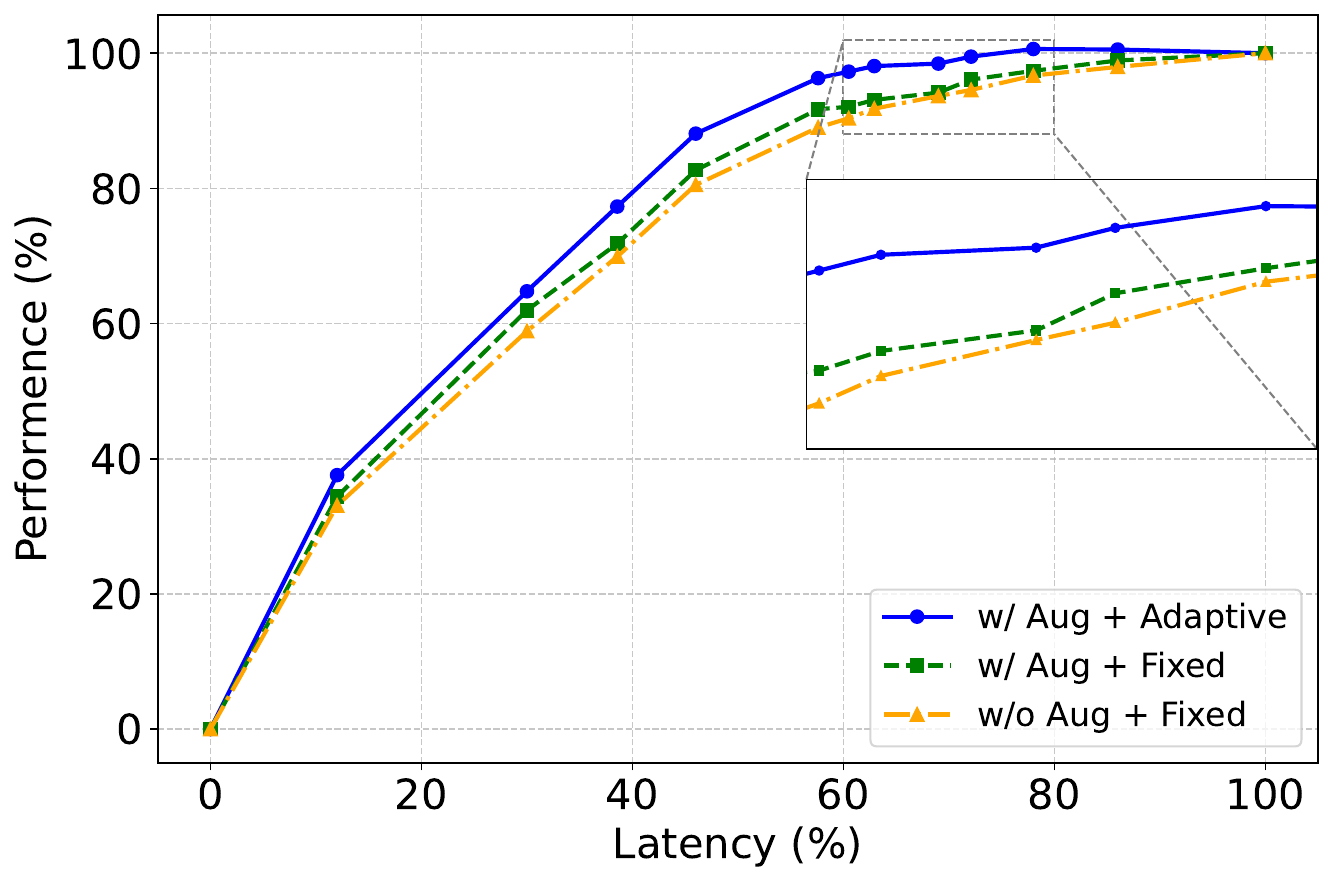}%
    \label{fig:ablation-latency}
}
\hfil
\subfloat[Performance vs Cost]{
    \includegraphics[width=0.88\linewidth]{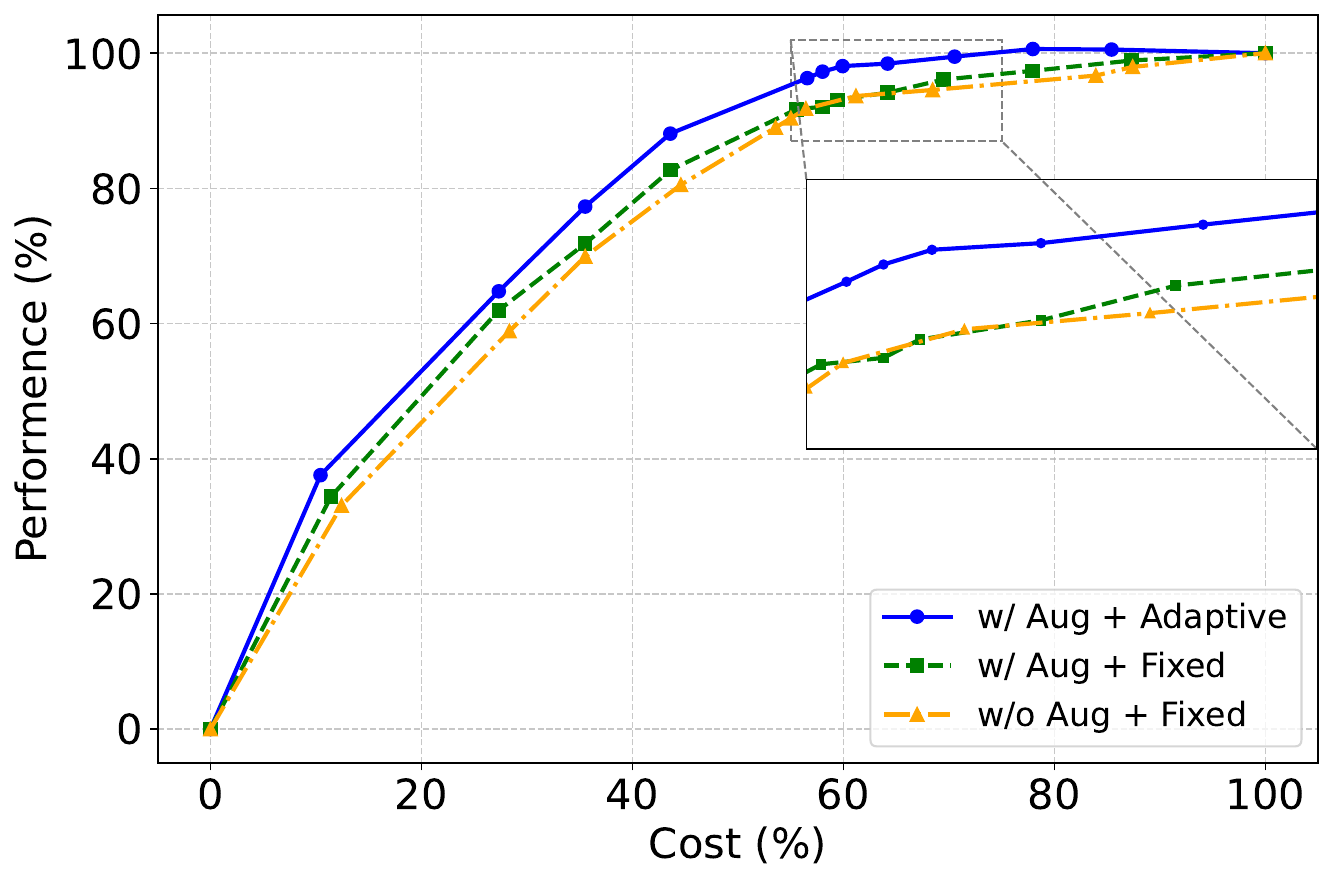}%
    \label{fig:ablation-cost}
}
\caption{Impact of supervision source on routing efficiency. Each curve shows the latency or cost required to reach a target CLM-only accuracy.}
\label{fig:ablation}
\end{figure}

\subsubsection{Effect of Label Augmentation}

Figure~\ref{fig:ablation} shows that incorporating label augmentation  improves routing accuracy. Compared to using reranker-only supervision, the addition of correctness- and consistency-based labels leads to a notable performance gain. This confirms that enriched supervision provides more reliable guidance for the consistency predictor.

\subsubsection{Effect of Dynamic Thresholding}

Figure~\ref{fig:ablation} demonstrates that using dynamically optimized thresholds yields clear performance improvements over static thresholding. This validates that query-specific thresholds, learned via Bayesian optimization, enable the router to better adapt to diverse query complexities and risk levels. In contrast, static heuristics often fail to generalize across tasks, leading to suboptimal routing decisions.

The full model, which combines both dynamic thresholding and label augmentation, consistently achieves the best results across all metrics, demonstrating the complementarity of these two components.

\section{Conclusion}

This paper presents a lightweight and semantics-aware query routing framework for collaborative LLM inference across cloud, edge, and device tiers. Our approach leverages semantic consistency between outputs for more reliable routing. By combining reranker-based supervision with consistency-oriented data augmentation, and reusing deep hidden states for query representation, the framework enables adaptive, low-overhead routing. A class-specific thresholding strategy via Bayesian optimization further improves query sensitivity. Experiments show strong performance–efficiency trade-offs.In future work, we plan to extend ConsRoute to broader application domains and richer adaptivity signals, and to study its behavior under larger-scale, real-world mobile deployments.

\bibliographystyle{IEEEtran}
\bibliography{ref}

\vfill

\end{document}